\documentclass[journal]{IEEEtran}

\ifCLASSINFOpdf
\else
\fi

\usepackage{algorithm}
\usepackage{algpseudocode}
\usepackage{graphicx}
\usepackage{subcaption}
\usepackage{booktabs}
\usepackage{enumitem}
\usepackage{amssymb}
\usepackage{multirow}
\usepackage[cmex10]{amsmath}
\usepackage{cite}
\usepackage{color}

\newcommand{\rev}[1]{\textcolor{black}{#1}}
\newcommand{\revision}[1]{\textcolor{black}{#1}}

\begin{document}
%
\title{Illumination invariant hyperspectral image unmixing based on a digital surface model}

\author{Tatsumi~Uezato~\IEEEmembership{Member,~IEEE,}
        Naoto~Yokoya,~\IEEEmembership{Member,~IEEE,}
        and~Wei~He,~\IEEEmembership{Member,~IEEE}}

\markboth{}%
{Shell \MakeLowercase{\textit{et al.}}: Bare Demo of IEEEtran.cls for Journals}
\maketitle

\begin{abstract}
Although many spectral unmixing models have been developed to address spectral variability caused by variable incident illuminations, the mechanism of the spectral variability is still unclear. This paper proposes an unmixing model, named illumination invariant spectral unmixing (IISU). IISU makes the first attempt to use the radiance hyperspectral data and a LiDAR-derived digital surface model (DSM) in order to physically explain variable illuminations and shadows in the unmixing framework. Incident angles, sky factors, visibility from the sun derived from the LiDAR-derived DSM support the explicit explanation of endmember variability in the unmixing process from radiance perspective. The proposed model was efficiently solved by a straightforward optimization procedure. The unmixing results showed that the other state-of-the-art unmixing models did not work well especially in the shaded pixels. On the other hand, the proposed model estimated more accurate abundances and shadow compensated reflectance than the existing models. 
\end{abstract}


%
\IEEEpeerreviewmaketitle

\section{Introduction}
\IEEEPARstart{H}{yperspectral} imagery measures reflected light (i.e., radiance) at different wavelengths for each pixel of the image. The radiance is converted into a reflectance spectrum in order to remove the effects of atmospheric gases or water absorption~\cite{Griff2003}. The reflectance spectrum can be modelled by a linear combination of reference spectra representing the surface materials \cite{Kesha2002}. A linear mixing model (LMM) or a nonlinear mixing model (NLMM) has been developed to model the relationship between reference spectra (\textit{endmembers}) and observed reflectance spectra \cite{Biouc2012}. LMM enables spatial distributions of materials to be quantitatively estimated. While many mixing models have been used for a variety of applications reviewed by \cite{Biouc2012}, there are still two major problems.

The first problem is caused by a variable topography present in a hyperspectral image (HSI). The variable topography induces variations in incident illumination or shadows \cite{Murph2012}. The illumination variations or shadows adversely degrade the performance of spectral unmixing because of spectral variability within each endmember class \cite{Uezat2016}. Many methods have been developed to consider the spectral variability, which are reviewed by \cite{Somer2011,Zare2014a}. The most common approach is to use a single shade endmember or a scaling factor to model the variations in illumination \cite{Rober1998,Nasci2005a,Winte1999}. However, it is reported that the illumination variations cannot be simply modeled by the single shade endmember or scaling factor~\cite{Uezat2016a}. As a result, multiple shade endmembers are used to model the the spectral variability in \cite{Uezat2016a,Fitzg2005}. Some studies have also extended LMM by incorporating additional variability terms in the mixing model \cite{Drume2016,Thouv2016}.

While the developed unmixing models may be more robust to the spectral variability, they oversimplify the mechanism of spectral variability and lack clear physical meanings. For example, the models did not describe how the light interacts within each pixel and why the spectral variability is caused. The models commonly confuse the sources of spectral variability caused by illumination variations or physical characteristics of materials \cite{Uezat2016b}. Although direct sunlight and diffused skylight need to be considered to model spectra affected by illumination variations or shadows \cite{Ramak2015,Frima2011}, there are few unmixing models that incorporate these information. The variable topography also causes multiple scatterings of light, leading to nonlinear interactions of endmembers \cite{Dobig2014,Heyle2014}. The proportion of multiple scatterings may significantly increase in shaded pixels because the direct sunlight is blocked \cite{Adeli2013}. Although some NLMMs have been developed to describe the physical interactions of endmembers \cite{Megan2014a}, the models did not describe the physical interactions together with diffused skylight in shadows. There is a strong need to develop a model that can explicitly incorporate the spectral variability caused by the above problems. 

The second problem is that most LMMs or NLMMs assume that reflectance data are available \textit{a priori} and ignore an atmospheric correction step. However, HSI is commonly recorded as digital numbers corrected to radiance data. A user need to convert the radiance data to reflectance data by applying an atmospheric correction method \cite{Griff2003}. 
\revision{The atmospheric correction step is the key to understanding how the endmember variability is caused in the unmixing process. Most spectral unmixing studies aim to model endmember variability using available reflectance data. The modelling of endmember variability has been done by mathematical optimization without considering the physical meaning. By considering the atmospheric correction step, endmember variability can be physically modeled in the unmixing model. However, there are few spectral unmixing models that consider both atmospheric and unmixing steps in a unified unmixing framework.}

The two long-standing problems limit the use of spectral unmixing for HSI affected by illumination variations including shadows. Although there is a need to develop a robust unmixing model that addresses the problems, it is difficult to model the physical interactions of sunlight or skylight using only HSI. A digital surface model (DSM) acquired from LiDAR data is robust to the illumination variations and can be used with HSI \cite{Brell2017,Ni2014}. DSM has been used as spatial regularization in LMM \cite{Uezat2018} or used as preprocessing before the selection of endmembers \cite{Feng2003}. However, the studies did not use DSM to model the light interactions in the mixing model. DSM has great potential to estimate the geometry of topography~\cite{Frima2011} and to be used to model the physical interactions of endmembers. This paper proposes a model that addresses the aforementioned problems using hyperspectral data and LiDAR-derived DSM. The contributions of this paper are threefold:
\begin{enumerate}
  \item it proposes an illumination invariant spectral unmixing model (IISU) that can incorporate an atmospheric correction step and directly unmix radiance data; 
  \item it describes the physical meaning of existing unmixing models using parameters derived from hyperspectral data and LiDAR-derived DSM;
  \item it shows why and how existing unmixing models does not work well in shaded pixels and provides comparison between the proposed model and the existing unmixing models.
\end{enumerate}
This paper is organized as follows. Section \ref{section2} describes a novel unmixing model and existing models, while highlighting how the novel model overcomes the problems caused by variable illuminations. Section \ref{section3} provides the IISU-based algorithm and its optimization procedure. Section \ref{section4} and Section \ref{section5} show experimental results of simulated and real data. Finally, conclusions are drawn in Section \ref{section6}.

\section{Illumination invariant spectral unmixing}\label{section2}
\subsection{Notations}
$x$ represents a scalar, $\mathbf{x}$ represents a vector, $\mathbf{X}$ represents a matrix, $x_{bp}$ shows that $x$ is dependent on both a band and a pixel, $x_p$ shows that $x$ is dependent on only a pixel, $x_b$ shows that $x$ is dependent on only a band.\\
\\
\resizebox{.9\columnwidth}{!}{
\begin{tabular}{lcl}
\hline
&Definition\\
\hline
$P$& Number of pixels\\
$B$& Number of bands\\
$K$& Number of endmember classes\\
$l_{bp}$& Radiance of the $b$th band at the $p$th pixel\\
$s^1_{b}$& Direct sunlight of the $b$th band\\
& affected by atmospheric transmittance\\
$s^2_{b}$& Diffused skylight of the $b$th band\\
$\tilde{s}^1_{bp}$& Indirect sunlight of the $b$th band at the $p$th pixel\\
$\tilde{s}^2_{bp}$& Indirect skylight of the $b$th band at the $p$th pixel\\
$f_{p}$& Sky factor at the $p$th pixel\\
$\theta_{p}$& Angle between the direction of the sun\\
& and the surface normal of the $p$th pixel\\
$v_{p}$& Visibility at the $p$th pixel\\
$r_{bp}$& Reflectance of the $b$th band at the $p$th pixel\\
$y_{bp}$& Apparent reflectance of the $b$th band at the $p$th pixel\\
$m_{bk}$&  Endmember of the $b$th band of the $k$th class\\
$a_{kp}$&  Fractional abundance of the $k$th class at the $p$th pixel\\
$e_{jp}$&  Coefficient of the $j$th class at the $p$th pixel\\
$\odot$&  Element-wise product\\
$\oslash$&  Element-wise division\\
$\Vert\cdot\Vert_2$& $l_2$ norm\\
\hline
\end{tabular}
}
\begin{figure}[h!]
        \begin{subfigure}[b]{0.5\columnwidth}
                \centering
                \includegraphics[width=.9\linewidth]{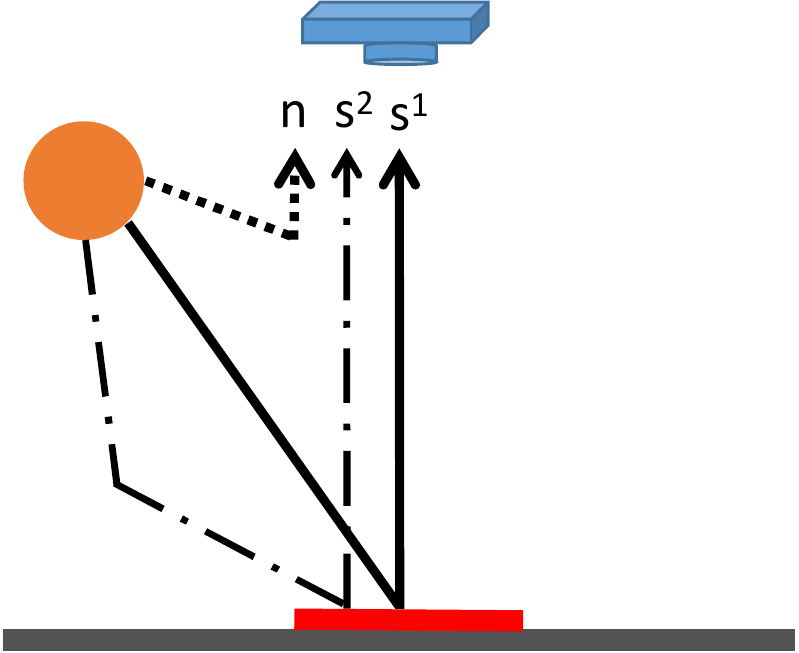}
                \caption{}
                \label{fig:light_1}
        \end{subfigure}%
        \begin{subfigure}[b]{0.5\columnwidth}
                \centering
                \includegraphics[width=.9\linewidth]{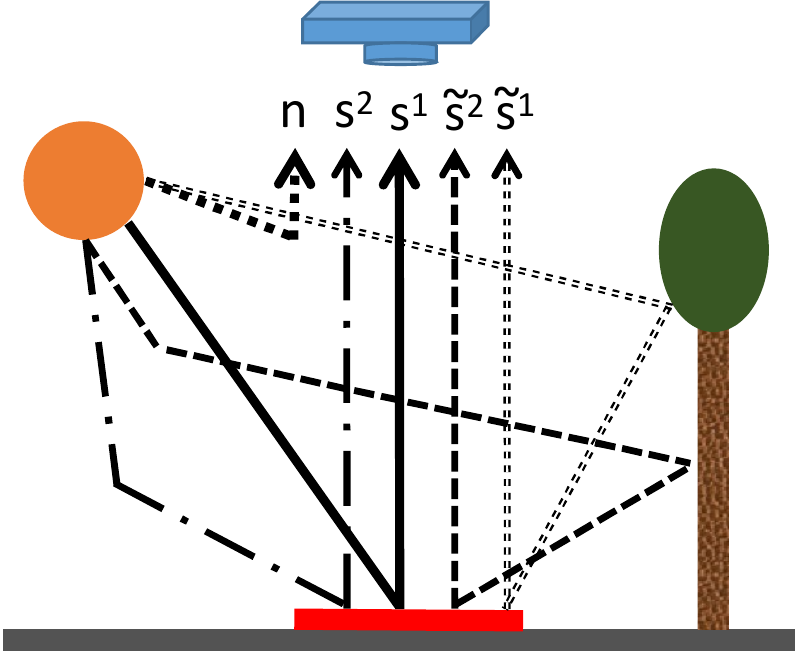}
                \caption{}
                \label{fig:light_2}
        \end{subfigure}
        \caption{Illustration of light path: (a) flat surface and (b) surface surrounded by an object. $n$ shows the path radiance. The red square represents a pixel.}\label{fig:rgb}
\end{figure}
\subsection{Problems of illumination variations}\label{sec:2b}
\textit{Apparent} reflectance spectra affected by the geometry and intensity of illumination are usually acquired by a calibration of HSI. In this section, a simple example that uses a calibration panel is given to describe why and how the apparent reflectance are acquired. Atmospheric correction is commonly used to derive reflectance spectra from at-sensor radiance. A simple atmospheric correction method can be done using a calibration panel. A simplified radiative transfer model is considered in this section  \cite{Ramak2015,Frima2011}. First, the at-sensor radiance derived from a calibration panel can be defined as follows:
\begin{equation}
	l_{b}=c_b\left[ s^1_b\cos\theta_c + fs^2_b \right],
\end{equation}
where $l_{b}$ is the at-sensor radiance of the $b$th band acquired from the calibration panel, $c_b$ is the reflectance of the calibration panel at the $b$th band, $s^1_b$ is the direct sunlight of the $b$th band, $\theta_c$ is the angle between the surface normal of the calibration panel and the direction of the sun. $s^2_b$ is the diffused skylight of the $b$th band, $f$ is the sky factor representing a amount of skylight falling on the surface. The calibration panel is assumed to be located in a place where there are no surrounding objects ($f=1$) and where the surface is flat. The path radiance is ignored for simplicity. In this model, we assume a Lambertian surface where the radiance reflected from a surface is uniform in any direction. If the at-sensor radiance is acquired from an object $o$ at the same condition with the calibration panel (Fig. \ref{fig:light_1}), the at-sensor radiance $l_{bo}$ derived from the object $o$ becomes:
\begin{equation}\label{eq:radiance_cpanel}
	l_{bo}=r_b\left[ s^1_b\cos\theta_o + fs^2_b \right],
\end{equation}
where $r_b$ is the reflectance of the $b$th band of the object, $\theta_o$ is the angle between the surface normal of the object and the direction of the sun. The reflectance of the object can be derived as follows:
\begin{equation}
	\frac{l_{bo}}{l_{b}}=\frac{r_b\left[ s^1_b\cos\theta_o + fs^2_b \right]}{c_b\left[ s^1_b\cos\theta_c + fs^2_b \right]}=r_b,
\end{equation}
where the reflectance $c_b$ of the calibration panel is assumed to be 1 and $\theta_o=\theta_c$ because the surface is flat. The reflectance is retrieved because the atmospheric effects are completely removed by the atmospheric correction step. However, each spectrum of a target pixel $p$ in HSI is usually acquired in a variable scene geometry. Indirect illumination also needs to be considered when the sunlight or skylight scattered by a different region reaches the target pixel $p$ (Fig. \ref{fig:light_2}).\footnote{\rev{The model considers a simplified linear and bilinear light interactions. However, it can potentially consider more complicated nonlinear interactions such as intimite mixtures of a tree canopy or minerals by increasing the number of parameters.}} The at-sensor radiance at the $p$th pixel can be:
\begin{equation}
	l_{bp}=r_{bp}\left[ v_ps^1_b\cos\theta_p + f_ps^2_b +\tilde{s}^1_{bp} +\tilde{s}^2_{bp}\right],
\end{equation}
where $r_{bp}$ is the reflectance of the $p$th pixel, $v_p$ is the binary value (1 or 0) where $v_p=0$ if the pixel is occluded from the sun and $v_p=1$ if the sunlight reaches the pixel, $f_p$ is the sky factor of the $p$th pixel, $\theta_p$ is the angle between the surface normal of the $p$th pixel and the direction of the sun, $\tilde{s}^1_{bp}$ is the scattered sunlight that reaches the $p$th pixel and $\tilde{s}^2_{bp}$ is the skylight scattered by an object that reaches the $p$th pixel. By using the calibration panel, the following equation is derived:
\begin{equation}\label{eq:apparent_r}
	y_{bp}=\frac{l_{bp}}{l_{b}}=\frac{\left[ v_ps^1_b\cos\theta_p + f_ps^2_b +\tilde{s}^1_{bp} +\tilde{s}^2_{bp}\right]}{\left[ s^1_b\cos\theta_c + fs^2_b \right]}r_{bp}.
\end{equation}
Note that $\theta_p$ is variable across all pixels while $\theta_c$ is fixed. The retrieved \textit{apparent} reflectance $y_{bp}$ is affected by a variety of factors (i.e., the viewing angle, the sky factor and the indirect illumination) and can be variable for each pixel \cite{Ramak2015}. These factors cause the following problems:
\begin{enumerate}
  \item \textbf{Variable viewing angle}: The variable sun angle changes the magnitude of the reflectance spectrum greatly. 
  \item \textbf{Variable sky factor}: When the sky factor is ignored in the atmospheric correction, it greatly changes the shape of the reflectance spectrum especially at shorter wavelengths (less than 550 nm). 
  \item \textbf{Indirect illumination}: The indirect illumination causes nonlinear effects that impact on both magnitude and shape of the reflectance spectrum.
\end{enumerate}
These factors cause spectral variability or nonlinearity for the apparent reflectance. Many existing unmixing models that incorporate spectral variability or nonlinearity aim to address the above problems. The example of this section only shows the atmospheric correction done by the use of a calibration panel. This is because it can show how the existing spectral unmixing methods can work using reflectance data derived from the standard approach. Note that there are also many sophisticated atmospheric correction methods and some methods can use DSM to consider shadows~(e.g. \cite{Cooley2002,Richter2004,Schlaepfer2018}). As shown in this section, an atmospheric correction step is important for spectral unmixing and can be useful to understand the causes of endmember variability in the unmixing models. However, the atmospheric correction and spectral unmixing methods have been studied independently. The importance of the atmospheric correction has been rarely considered in the unmixing models. \revision{This paper aims to propose a novel spectral unmixing model that considers all of the aforementioned factors with the help of LiDAR-derived DSM. The proposed model allows the detection of illumination variations and the definition of specific endmembers permitting their unmixing.}
%
%
\subsection{Illumination invariant spectral unmixing (IISU)}
The proposed illumination invariant spectral unmixing model (IISU) is based on a simplified radiative transfer model as follows:
\begin{equation}\label{eq:iisu_s}
	l_{bp}=r_{bp}\left[ v_ps^1_b\cos\theta_p + f_ps^2_b + \tilde{s}^1_{bp} \right] + n_{bp},
\end{equation}
where $n_{bp}$ is an additive noise of the $b$th band at the $p$th pixel. Note that the path radiance is ignored for simplicity. The diffused skylight and noise term (i.e., $f_p$, $s^2_b$ and $n_{bp}$) can compensate the contribution of the path radiance. $\tilde{s}^2_p$ is not considered in this model because it can be negligible compared with other light sources~\cite{Adeli2013}. The shadow compensated reflectance can be modeled using endmembers as
\begin{equation}\label{eq:ir}
	r_{bp}=\sum_{k=1}^{K}m_{bk}a_{kp}
\end{equation}
where $m_{bk}$ is the reflectance of the $b$th band of the $k$th endmember, $a_{kp}$ is the $k$th abundance fraction at the $p$th pixel. The indirect sunlight can also be modeled using endmembers as
\begin{equation}\label{eq:ts}
	\tilde{s}^1_{bp}=\sum_{j=1}^{K}s^1_b m_{bj}e_{jp},
\end{equation}
where $e_{jp}$ is the coefficient ($\geq 0$) representing the contribution of indirect sunlight scattered by the $j$th endmember. The indirect sunlight is modeled by the sunlight reflected by different endmembers and its contribution is controlled by $e$. By substituting (\ref{eq:ir}) and (\ref{eq:ts}) into (\ref{eq:iisu_s}), IISU models the at-sensor radiance of the $b$th band at the $p$th pixel as follows:
\begin{equation}
	l_{bp}=\sum_{k=1}^{K}m_{bk}a_{kp}\Bigl[ v_ps^1_b\cos\theta_p + f_ps^2_b +\sum_{j=1}^{K}s^1_b m_{bj}e_{jp}\Bigr] + n_{bp}.
\end{equation}
When considering all bands representing wavelengths, $\mathbf{l}_p \in  \mathbb{R}^{B \times 1}$ is equivalent to:
\begin{equation}\label{eq:iisu}
    \begin{aligned}
	\mathbf{l}_p=\sum_{k=1}^{K}a_{kp} \Bigl[\mathbf{m}_k \odot \Big\{ &\mathbf{s}^1 v_p\cos\theta_p \\
	&+ \mathbf{s}^2f_p +\sum_{j=1}^{K}e_{jp} \mathbf{m}_j \odot \mathbf{s}^1 \Big\} \Bigr] + \mathbf{n}_p,\\
	\forall k,\forall p,\  &a_{kp}\geq 0, \quad \sum_{k=1}^{K}a_{kp}=1.
	\end{aligned}
\end{equation}
The novelty of this model is i) to naturally incorporate parameters ($f$, $\theta$, $v$) derived from LiDAR-derived DSM in the unmixing model; ii) to explicitly model light interactions under various illumination conditions; iii) to directly fit the unmixing model using the radiance, not apparent reflectance. The model enables variable factors estimated from LiDAR-derived DSM to be considered for spectral variability within each pixel. Once the abundances are estimated, IISU can recover the shadow compensated reflectance as
\begin{equation}
	\mathbf{r}_p=\sum_{k=1}^{K}\mathbf{m}_ka_{kp}.
\end{equation}
This shows that IISU can \textit{unmix} spectral mixtures caused by skylight or indirect sunlight and recover the shadow compensated reflectance.
\subsection{Related unmixing models}\label{relatedwork}
In this section, we show that there is a close relationship between existing unmixing models and IISU. It is possible to describe the physical meaning of the existing models from the perspective of the proposed physical model. Existing models usually unmix the apparent reflectance as described in the section \ref{sec:2b}. The various factors that cause spectral variability are propagated and included in the apparent reflectance spectrum $\mathbf{y}_p$. From (\ref{eq:apparent_r}), (\ref{eq:ir}) and (\ref{eq:ts}), the following equation can be derived
\begin{equation}
\begin{aligned}
	\mathbf{y}_p=\sum_{k=1}^{K}a_{kp}\Bigl[ & \left(\mathbf{m}_k \odot \mathbf{w}^1 \right) v_p\cos\theta_p + \left(\mathbf{m}_k \odot \mathbf{w}^2 \right)f_p \\
	&+\sum_{j=1}^{K}e_{jp}^1\left(\mathbf{m}_k \odot \mathbf{m}_j \odot \mathbf{w}^1 \right) \Bigr],
	\end{aligned}
\end{equation}
where $\mathbf{w}^1=\mathbf{s}^1 \oslash (\mathbf{s}^1\cos\theta_c + \mathbf{s}^2)$ and $\mathbf{w}^2=\mathbf{s}^2 \oslash (\mathbf{s}^1\cos\theta_c + s^2)$ represent the contribution of sunlight and skylight, respectively. Noise is ignored for simplicity. This model generalizes many existing unmixing models and can describe the physical meaning of the models.

When ignoring skylight ($\mathbf{s}^2=0$) and indirect sunlight ($e_{jp}=0$) and all pixels are visible from the sun ($v_p=1$), the model can be written as
\begin{equation}
	\mathbf{y}_p=\sum_{k=1}^{K}\mathbf{m}_k \tau_pa_{kp},
\end{equation}
where $\tau_p=\frac{\cos\theta_p}{\cos\theta_c}$ and the variable topography causes a scaling factor. This model is equivalent to LMM with a scaling factor~\cite{Nasci2005a}. The scaling factor can also be incorporated by using a shade endmember.

When the indirect sunlight is ignored ($e_{jp}=0$),
\begin{equation}\label{eq:model2}
	\mathbf{y}_p=\sum_{k=1}^{K}a_{kp}\left[\mathbf{m}_k\odot \left( \mathbf{w}^1 v_p\cos\theta_p + \mathbf{w}^2f_p \right)\right].
\end{equation}
This model shows that each endmember spectrum is affected by the skylight, leading to endmember variability. Endmember variability caused by skylight is also considered in \cite{Uezat2016a}.

When the skylight is ignored and $\tau_p$ is fixed as 1 ($\mathbf{s}^2=0,~\tau_p=1$), $\mathbf{y}_p$ can be modeled as:
\begin{equation}\label{eq:model3}
	\mathbf{y}_p=\sum_{k=1}^{K}\mathbf{m}_k a_{kp}  + \sum_{k=1}^{K}\sum_{j=1}^{K}b_{kj}\left(\mathbf{m}_k \odot \mathbf{m}_j\right),
\end{equation}
where $b_{kj}=\frac{e_{jp}a_{kp}}{\cos\theta_c}$. This model can be interpreted as the bilinear model that has been widely studied in \cite{Altma2012,Dobig2014,Heyle2014,Megan2014a,Yokoy2014}. This shows that these models can incorporate the indirect sunlight scattered by endmembers. Among them, IISU is most similar to the bilinear model developed in \cite{Megan2014,qu2014}.

When considering all parameters including the skylight and the indirect sunlight,
\begin{equation}\label{eq:model4}
\begin{aligned}
	\mathbf{y}_p&=\sum_{k=1}^{K}\left[\mathbf{m}_k\odot \left( \mathbf{w}^1 v_p\cos\theta_p + \mathbf{w}^2f_p +\mathbf{M}\mathbf{e}_p \right)\right]a_{kp}\\
	&=\mathbf{M}_p\mathbf{a}_p,
	\end{aligned}
\end{equation}
where $\mathbf{M}_p=\text{diag}\left( \mathbf{w}^1 v_p\cos\theta_p + \mathbf{w}^2f_p +\mathbf{M}\mathbf{e}_p \right)\mathbf{M}$. This shows that pixel-wise endmember variability occurs for each band while the amount is fixed among all endmember classes. The equation helps us understand the physical meaning of models that incorporate endmember variability. For example, the model in \cite{Drume2016} considers endmember variability as a scaling factor. 
\begin{equation}
	\mathbf{y}_p= \mathbf{M}\mathbf{Z}_p \mathbf{a}_{p},
\end{equation}
where $\mathbf{Z}_p \in \mathbb{R}^{K \times K}$ is a diagonal ma
trix where each diagonal element shows endmember variability. This shows that the model simplifies the band-dependent endmember variability as a scalar value across all bands. This may happen when the contributions of skylight and indirect sunlight are negligible. The model in \cite{Thouv2016} incorporates endmember variability as additive factors.
\begin{equation}
	\mathbf{y}_p= (\mathbf{M}+\mathbf{dM}) \mathbf{a}_{p},
\end{equation}
where $\mathbf{dM}$ represents an additive factor. This implies that there is no spectral variability caused by topography and incorporates the indirect sunlight or skylight terms as the additive factor. The model incorporating both multiplicative and additive terms has been also developed by \cite{Liu2017} as 
\begin{equation}
	\mathbf{y}_p = \mathbf{C}_p\mathbf{M}\mathbf{a}_{p} + \Phi_p,
\end{equation}
where $\mathbf{C}_p \in \mathbb{R}^{B \times B}$ is a diagonal matrix where each diagonal element shows a band-dependent endmember variability, $\Phi_p$ is considered as an additive term. Note that the notations are simplified from \cite{Liu2017}. This model can incorporate the spectral variability caused by variable topography as $\mathbf{C}_p$ and spectral variability caused by skylight or indirect sunlight as $\Phi$. Similarly, the model developed by~\cite{Hong2019} assumes
\begin{equation}
	\mathbf{y}_p = \mathbf{M}\tau_p \mathbf{a}_{p} + \Phi_p,
\end{equation}
where the endmember variability is simplified as a scaling factor $\tau_p$ while incorporating the additive factor. The model developed by \cite{Imbir2017} can incorporate band-dependent spectral variability as
\begin{equation}
	\mathbf{y}_p=(\mathbf{M} \odot \mathbf{C}_p) \mathbf{a}_{p},
\end{equation}
where $\mathbf{C}_p \in \mathbb{R}^{B \times K}$ represents the band and class dependent spectral variability. This model is most similar to (\ref{eq:model4}) and can describe the spectral variability caused by the various factors as $\mathbf{C}_p$.

Other state-of-the-art spectral unmixing models incorporate the spectral variability using multiple endmember spectra within each class \cite{Uezat2019,Uezat2016,Drume2019}. The models use multiple endmember spectra affected by the variable illuminations and simplify the spectral variability as a convex hull or cone spanned by the endmember spectra. There are also methods that use a probability distribution to model the spectral variability \cite{Zhou2018,Halim2016}.

As shown above, IISU shows that the existing spectral unmixing models can incorporate the part of spectral variability from the physical perspective of the light interaction. However, it also shows two major problems. First, the existing models may fail to unmix a pixel spectrum in shadows where $v_p=0$. When a pixel is in a shadow, the pixel-wise endmembers need to be modeled by only skylight and indirect sunlight. However, the existing models cannot consider the pixel-wise endmembers. Second, it is challenging to optimize the parameters representing endmember variability using only HSI. This is because the endmember variability is caused by various factors including shadows and can be very large. Although the parameters can be simplified to ease the optimization, the simplified parameters may lose important information about the endmember variability. \rev{IISU is different to the existing spectral unmixing models in that i) it can use parameters derived from LiDAR-derived DSM to model the endmember variability in the unmixing process: ii) its optimization can be straightforward while incorporating flexible endmember variability. Compared with the spectral unmixing models that require complicated hyper-parameter tuning, IISU optimizes fewer number of parameters with the help of LiDAR-derived DSM as shown in the next section.}
\section{IISU-based unmixing algorithm}\label{section3}
In (\ref{eq:iisu}), the parameters of $v$, $f$ and $\theta$ can be estimated from LiDAR-derived DSM as shown in the supplementary material. There are still 4 unknown parameters ($a$, $e$, $\mathbf{s}^1$ and $\mathbf{s}^2$). Simultaneous estimation of these parameters is challenging. IISU firstly estimate $\mathbf{s}^1$ and $\mathbf{s}^2$ in a preprocessing step and then estimate $a$ and $e$ in a main step. By taking the 2 steps, the optimization can be more efficient and easier to be solved. In this section, two optimization problems are proposed and solved in the preprocessing and main steps.
\subsection{Preprocessing}
In order to estimate abundances in (\ref{eq:iisu}), the parameters ($\mathbf{s}^1$ and $\mathbf{s}^2$) representing sunlight and skylight need to be provided. This preprocessing step is designed to estimate the two parameters. First, we manually find $N$ pixels representing a same object in sunlit and in shadow\footnote{Only one object is enough for the preprocessing step. However, the use of multiple objects can lead to better estimates of sunlight and skylight if available.}. This can be done by using prior knowledge or looking at the RGB image. When assuming the reference endmember spectrum (i.e., $\mathbf{m}$) of the object is available, (\ref{eq:iisu}) can be rewritten as
\begin{equation}\label{eq:esti_s1}
    \begin{aligned}
	\mathbf{L}=(\mathbf{m}\mathbf{1}^T) \odot \Bigl[ \mathbf{s}^1 &(\mathbf{v}\odot\cos\boldsymbol\theta)^T + \mathbf{s}^2\mathbf{f}^T\Bigr] \\
	&+  (\mathbf{B}\mathbf{E}) \odot (\mathbf{s}^1\mathbf{1}^T) + \mathbf{N},
	\end{aligned}
\end{equation}
where $\mathbf{L} \in \mathbb{R}^{B \times N}$ is the radiance of the $N$ selected pixels in sunlit and in shadow, $\mathbf{1} \in \mathbb{R}^{N \times 1}$ is a vector where all elements are 1, $\mathbf{v}\odot\cos\boldsymbol\theta \in \mathbb{R}^{N \times 1}$ is a vector where each element represents $v_p\cos\theta_p$ and $\mathbf{f} \in \mathbb{R}^{N \times 1}$ are the sky factors of the selected pixels, $\mathbf{B}=\mathbf{M} \odot (\mathbf{m}\mathbf{1}^T)$ represents bilinear endmembers, $\mathbf{E}=[\mathbf{e}_1\mid ...\mid \mathbf{e}_N] \in \mathbb{R}^{K \times N}$ is the coefficients of the bilinear endmembers at the selected pixels. Although the optimization problem is non-convex, it can be convex with respect to each block of variables ($\mathbf{s}$ and $\mathbf{E}$). The optimization problem can be solved by using the block coordinate descent. The variables are estimated by alternatively solving the following two minimization problems. The minimization problem of $\mathbf{s}^1$ and $\mathbf{s}^2$ can be formulated as follows
\begin{eqnarray}\label{eq:esti_s}
    \begin{aligned}
	& \ \min\limits_{\mathbf{s}}\frac{1}{2}\Vert\mathbf{l}-\mathbf{T}\mathbf{s}\Vert^2_2, \\
     &\text{subject to } \mathbf{s}\succeq \mathbf{0},
    \end{aligned}
\end{eqnarray}
where $\mathbf{T}=\begin{bmatrix}\mathbf{T}^1 & \mathbf{T}^2\end{bmatrix}$, $\mathbf{s}=\begin{bmatrix}
\mathbf{s}^1 \\ \mathbf{s}^2 \end{bmatrix}$, $\mathbf{l}=\text{vec}(\mathbf{L})$, $\text{vec}(\cdot)$ represents the vectorization of a matrix, $\succeq$ is element-wise comparison. $\mathbf{T}^1$ and $\mathbf{T}^2$ are defined as
\begin{equation}
    \begin{aligned}
	&\mathbf{T}^1=\\
	&\begin{bmatrix} \text{diag}(\mathbf{m}v_1\cos\theta_1 + \mathbf{B}\mathbf{e}_1) \mid ... \mid \text{diag}(\mathbf{m}v_N\cos\theta_N + \mathbf{B}\mathbf{e}_N)\end{bmatrix}^T,
	\end{aligned}
\end{equation}
\begin{equation}
	\mathbf{T}^2=\begin{bmatrix} \text{diag}(\mathbf{m}f_1) \mid ... \mid \text{diag}(\mathbf{m}f_N)\end{bmatrix}^T.
\end{equation}
This is solved by the nonnegative least squares~\cite{Heinz2001}. The minimization problem of $\mathbf{E}$ is as follows:
\begin{eqnarray}\label{eq:esti_e}
	\begin{aligned}
	\min\limits_{\mathbf{E}}&\frac{1}{2}\Vert\mathbf{G}-\mathbf{\tilde{B}}\mathbf{E})\Vert^2_2 , \ \text{subject to } \mathbf{E}\succeq \mathbf{0},
    \end{aligned}
\end{eqnarray}
where $\mathbf{G}=\mathbf{L}-(\mathbf{m}\mathbf{1}^T) \odot \left[ \mathbf{s}^1 (\mathbf{v}\odot\cos\boldsymbol\theta)^T + \mathbf{s}^2\mathbf{f}^T\right]$, $\mathbf{\tilde{B}}=\text{diag}(\mathbf{s}^1)\mathbf{B}$. The two variables ($\mathbf{s}$ and $\mathbf{E}$) can be solved by iterating the nonnegative least squares until it converges.
\subsection{Estimation of abundances}
The model (\ref{eq:iisu}) is equivalent to
\begin{eqnarray}
\begin{aligned}
	\mathbf{l}_p=&\text{diag}(\mathbf{s}^1 v_p\cos\theta_p + \mathbf{s}^2f_p)\mathbf{M}\mathbf{a}_p +\\ &\sum_{k=1}^{K}\sum_{j=1}^{K}a_{kp}e_{jp}\left(\mathbf{m}_k \odot \mathbf{m}_j \odot \mathbf{s}^1\right) + \mathbf{n}_p.
	\end{aligned}
\end{eqnarray}
The optimization problem is non-convex because of the term $a_{kp}e_{jp}$. By replacing the term with a new variable $x$, the model can be rewritten as
\begin{eqnarray*}
\begin{aligned}
	\mathbf{l}_p=\text{diag}(\mathbf{s}^1 v_p\cos\theta_p + \mathbf{s}^2f_p)\mathbf{M}\mathbf{a}_p +\text{diag}(\mathbf{s}^1)\Xi\mathbf{x}_p + \mathbf{n}_p,\\
	\end{aligned}
\end{eqnarray*}
where $\Xi=\begin{bmatrix}\mathbf{m}_1 \odot \mathbf{m}_1 \mid \mathbf{m}_1 \odot \mathbf{m}_2 \mid ... \mid \mathbf{m}_K \odot \mathbf{m}_K\end{bmatrix} \in \mathbb{R}^{B \times R}$ represents bilinear endmembers, $R=\frac{1}{2}K(K +1)$, $\mathbf{x}_p \in \mathbb{R}^{R \times 1}$ shows the coefficients of the bilinear endmembers. The minimization problem of this model can be written as
\begin{equation}\label{eq:esti_a}
	\min\limits_{\mathbf{\tilde{a}}_p}\frac{1}{2}\Vert\mathbf{l}_p-\mathbf{\tilde{M}}_p\mathbf{\tilde{a}}_p \Vert_2^2, \ \text{subject to } \mathbf{\tilde{a}}_p\succeq \mathbf{0},
\end{equation}
where $\mathbf{\tilde{M}}_p=\begin{bmatrix} \text{diag}(\mathbf{s}^1 v_p\cos\theta_p + \mathbf{s}^2f_p)\mathbf{M} & \text{diag}(\mathbf{s}^1)\Xi\end{bmatrix}$, $\mathbf{\tilde{a}}_p=\begin{bmatrix}
\mathbf{a}_p \\ \mathbf{x}_p \end{bmatrix}$. $\mathbf{\tilde{M}}_p$ may show high collinearity among the bilinear endmembers. The alternating direction method of multipliers (ADMM;~\cite{Boyd2011}) is used to solve the problem because it can be robust to the collinearity problem~\cite{Sun2014}. Note that ASC is not considered because the estimate of $\mathbf{s}$ may be spatially variable and ASC may adversely affect the performance of unmixing. {The estimated $\mathbf{a}_p$ is normalized in order to address the spectral variability caused by physical characteristics of materials and obtain $\mathbf{a}_p$ that satisfies the ASC.} The other constraints are not considered in this study in order to test whether the proposed model can perform well without relying on the constraints. Note that other constraints (e.g. sparsity or spatial regularization) can be easily incorporated in the proposed optimization framework.
\begin{algorithm}
\caption{Algorithm for IISU-based unmixing}\label{algorithm}
\begin{algorithmic}[1]
\State \textbf{Step 1:} \text{Preprocessing}
\State $\mathbf{Input}: v_p, f_p, \theta$, and $\mathbf{L}$ of selected pixels in sunlit and shadow
\State $\mathbf{E}^{(0)}=\mathbf{0}$
\While{the stopping criterion is not satisfied}
\State estimate $\mathbf{s}^{(t+1)}$ by solving (\ref{eq:esti_s}).
\State estimate $\mathbf{E}^{(t+1)}$ by computing (\ref{eq:esti_e}).
\EndWhile
\State $\mathbf{Output}: \mathbf{s}_1^{(t+1)},\mathbf{s}_2^{(t+1)}$
\\
\State \textbf{Step 2:} \text{Estimation of abundances}
\State $\mathbf{Input}:\mathbf{s}_1^{(t+1)}, \mathbf{s}_2^{(t+1)}, \mathbf{M}, \mathbf{l}_p$
\For{$p \gets 1$ to $P$}
\State estimate $\mathbf{\tilde{a}}_p$ by solving (\ref{eq:esti_a}).
\State extract and normalize $\mathbf{a}_p$ from $\mathbf{\tilde{a}}_p$
\EndFor
\State $\mathbf{Output}: \mathbf{A} = \left[\mathbf{a}_1,\ldots,\mathbf{a}_P\right]$, \ $\mathbf{R}=\mathbf{M}\mathbf{A}$
\end{algorithmic}
\end{algorithm}
\subsection{Algorithmic scheme}
The general algorithmic scheme of IISU is detailed in Algorithm \ref{algorithm}. IISU comprises the simple two steps. In the first step, it estimates $\mathbf{s}_1$ and $\mathbf{s}_2$ using the block coordinate descend. The iteration stops when the change of updated values in $\frac{1}{2}\Vert(\mathbf{l}-\mathbf{T}\mathbf{s}-\mathbf{b})\Vert^2_2$ is smaller than $10^{-4}$. The second step estimates $\mathbf{A}$ by solving the convex optimization problem. Finally, the shadow compensated reflectance $\mathbf{R}$ is recovered by multiplying the endmembers by the estimated abundances.
\begin{figure}[h!]
        \begin{subfigure}[b]{0.5\columnwidth}
                \centering
                \includegraphics[width=.9\linewidth]{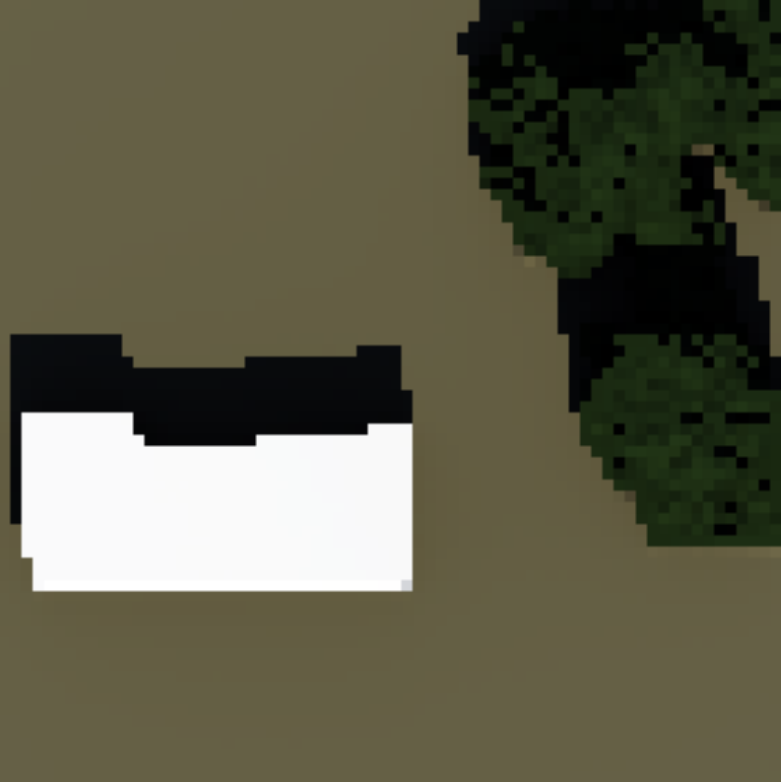}
                \caption{}
                \label{fig:sim1_rgb}
        \end{subfigure}%
        \begin{subfigure}[b]{0.5\columnwidth}
                \centering
                \includegraphics[width=.9\linewidth]{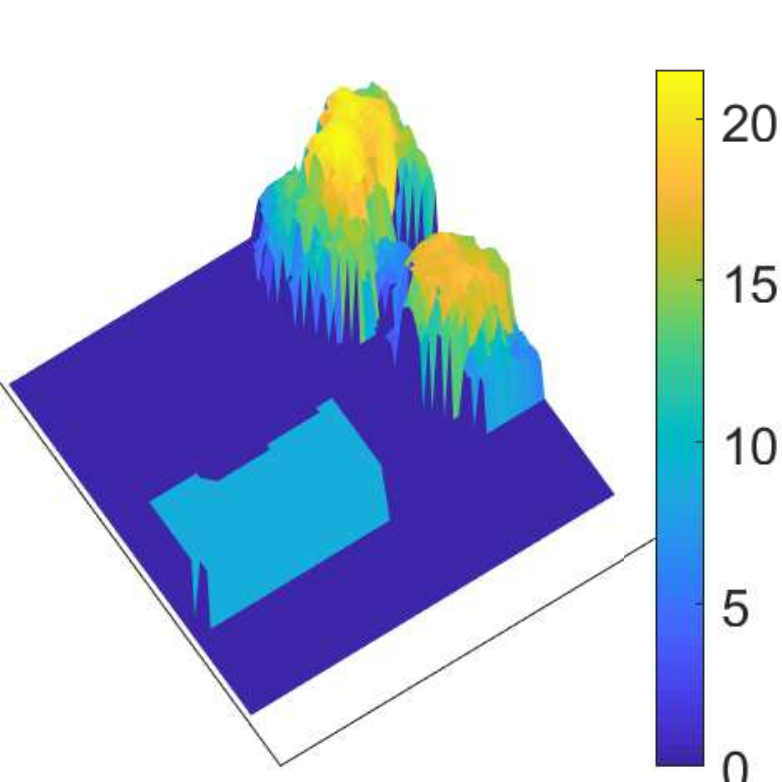}
                \caption{}
                \label{fig:sim1_dsm}
        \end{subfigure}
        \caption{SIM1: (a) RGB of the synthetic HSI. (b) DSM.}\label{fig:sim1}
\end{figure}
\begin{figure}[h!]
        \begin{subfigure}[b]{0.5\columnwidth}
                \centering
                \includegraphics[width=.9\linewidth]{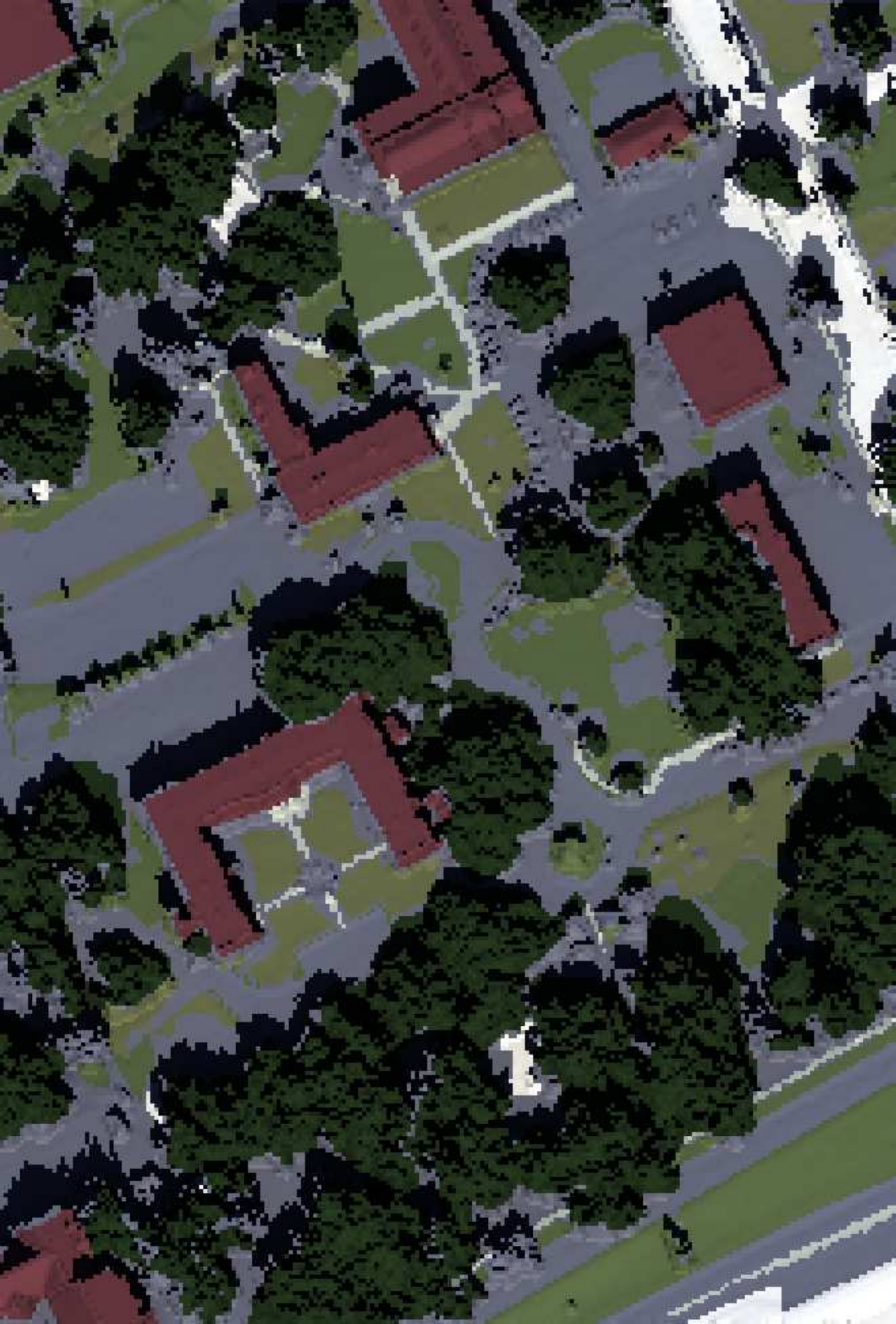}
                \caption{}
                \label{fig:sim2_rgb}
        \end{subfigure}%
        \begin{subfigure}[b]{0.5\columnwidth}
                \centering
                \includegraphics[width=.9\linewidth]{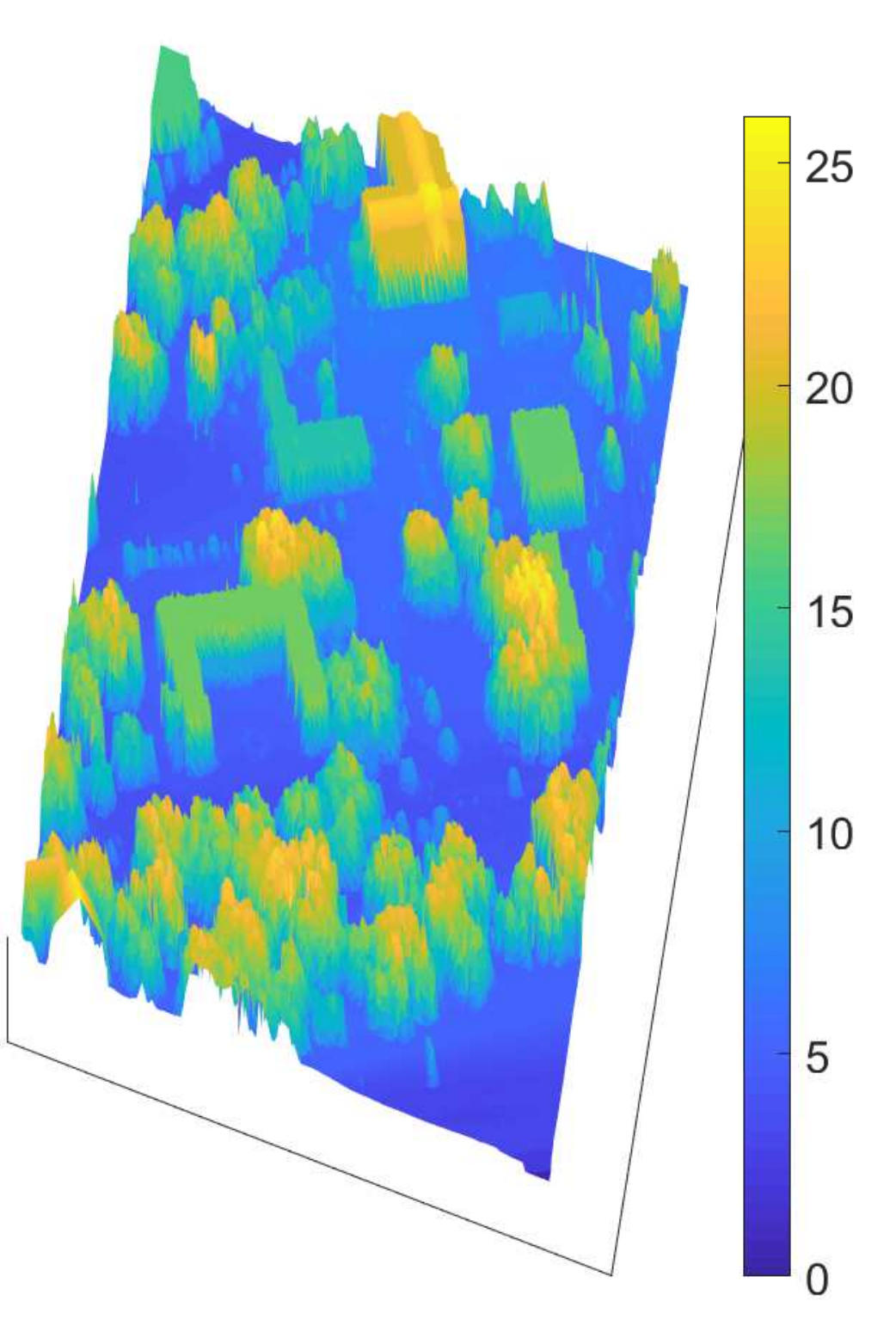}
                \caption{}
                \label{fig:sim2_dsm}
        \end{subfigure}
        \caption{SIM2: (a) RGB of the synthetic HSI. (b) DSM.}\label{fig:sim2}
\end{figure}
\section{Experiments using simulated data}\label{section4}
Simulated data are generated and used to demonstrate why existing methods fail to work in shaded pixels and how the proposed method outperforms other existing unmixing models. In the simulated data, the zenith and azimuth angles are assumed to be 40 and 190, respectively. The simulated data also assume that very high resolution (VHR) images (i.e., $1$~m) are acquired because variable incident illumination and shadows become large. Because of the VHR image, each pixel represents a pure pixel. The objective of this experiment is not to estimate fractional abundances but to investigate whether the models can unmix spectral mixtures caused by variable incident illuminations and retrieve shadow compensated reflectance spectra. In order to test and show the efficacy of the spectral unmixing methods that address the spectral variability, a standard atmospheric correction method is used in this experiment before the compared unmixing methods are applied.
%
\subsection{Simulated datasets}
\subsubsection{Simulated dataset 1 (SIM1)}
Simple homogeneous regions are considered in SIM1 in order to demonstrate how the existing models fail to work in shaded pixels. SIM1 is composed of three spatially homogeneous regions of trees, a soil, and a building. SIM1 was generated as follows. First, DSM (70 x 70) showing different heights for different materials was generated by extracting and smoothing a subset of DSM used in \cite{Uezat2018}. Second, 3 endmembers of a tree, a soil and a building was manually selected from HSI~\cite{Gader2013}. Third, DSM was manually segmented and abundance maps were generated from the segmentation map. Each abundance map of a material represents a binary map. Finally, a synthetic HSI was generated from the equation (\ref{eq:iisu}) by using the 3 endmembers, the abundance maps, the parameters ($v$,$f$,$\theta$) derived from DSM and the sunlight and skylight ($s^1$ and $s^2$) simulated from the simple model of the atmospheric radiative transfer of sunshine (SMARTS~\cite{Gueym1995}). In (\ref{eq:iisu}), the value of the coefficient $e$ was set as $0.01$ in shaded pixels because the contribution of indirect sunlight is usually small. The different values of $e$ were tested and shown in the supplementary document. Finally, a high signal to noise ratio (SNR= 50 dB) was considered in this main document in order to illustrate the errors that were mainly caused by the variable incident illumination and shadows. In the supplementary material, different levels of noise (i.e., $50$, $40$, and $30$ dB) were considered and the results were shown.
\subsubsection{Simulated dataset 2 (SIM2)}
SIM2 considers a more realistic scenario. A real DSM acquired in~\cite{Gader2013} was used with a synthetic image (325 x 220). The synthetic image was generated as follows. First, 7 endmember classes (tree, grass, ground surface, sand, road, building and sidewalk) were selected by using the ground truth provided by \cite{du2017}. The 7 endmember classes were selected because spectra of some classes in ground truth do not show the difference. The mean of spectra within each class was used as an endmember spectrum of each class. Second, the abundance maps of the selected endmember classes were generated by using ground truth. The binary abundance maps represent 1 in pixels where each endmember class is present while they represent 0 in pixels where the endmember is not present. Finally, the synthetic HSI was generated from the equation (\ref{eq:iisu}) by using the 7 endmembers, the abundances, the parameters ($v$,$f$,$\theta$) derived from the real DSM and the sunlight and skylight simulated from SMARTS. Like the simulation procedure of SIM1, the value of the coefficient $e$ was set as $0.01$ in shaded pixels. And SNR was set as 50~dB to illustrate the errors caused by the variable incident illuminations in this main document. However, the different values of $e$ and levels of noise were shown in the supplementary material.
\subsection{Compared methods}\label{compared_method}
Some of the related models discussed in Section \ref{relatedwork} and the state-of-the-art methods have been compared in this work.
\subsubsection{FCLS~\cite{Heinz2001}}
The traditional fully constrained least squares (FCLS) solves the unmixing problem with ASC and ANC. FCLS was used for comparison to show how much errors are caused when ignoring spectral variability.
\subsubsection{FCLS-s}
FCLS-s represents FCLS with a shade endmember. The reflectance values of the shade endmember are 0.001 across all bands. The shade endmember has been most widely used to address variable incident illuminations. FCLS-s was used as a baseline method.
\subsubsection{SCLS~\cite{Drume2016}}
The scaled constrained least square (SCLS) assumes that spectral variability is a scaling factor. SCLS was included for comparison to show the difference between spectral variability representing a simple scaling factor and spectral variability incorporating various factors derived from LiDAR-derived DSM.
\subsubsection{NLMM~\cite{Nasci2009}}
The nonlinear mixing model (NLMM) that incorporates bilinear endmembers in a similar way with IISU was compared to show the performance difference between a NLMM-based method and IISU. Note that a shade endmember was used to represent the scaling spectral variability for fair comparison.
\subsubsection{U-DSM~\cite{Uezat2018}}
The state-of-the-art unmixing method that incorporates DSM (U-DSM) was used for comparison. The method, called w-DSM in~\cite{Uezat2018}, uses DSM to promote spatial smoothness in estimated abunance maps. The method was included to show the performance difference between the method incorporating DSM and IISU. 
\subsection{Initialization}
All methods except IISU require reflectance data. In this study, radiance data at a white calibration panel (100$\%$ reflection across all bands) were assumed to be available. Radiance data of the HSIs were converted to reflectance data using the calibration panel as described in section \ref{sec:2b}. Only IISU used radiance data. Endmember spectra were assumed to be available \textit{a priori}. The parameter required for U-DSM was empirically determined in the set $(0.0001, 0.001, 0.01, 0.1, 1, 10)$ for each dataset so that the selected value produced the lowest RMSE$_a$. The parameters ($v$,$f$,$\theta$) required for IISU were estimated from each DSM. Twenty-five pixels of a soil class in sunlit and shaded areas were extracted for IISU by examining the RGB images. 
\subsection{Performance criterion}
Results of the methods were quantitatively evaluated using 3 different criteria. The root mean square error (RMSE) has been used to evaluate the estimated abundances and recovered shadow compensated reflectance. RMSE between true and estimated abundances has been defined as
\begin{equation}
\text{RMSE}_a =\sqrt{\frac{1}{P K}\sum\limits_{p = 1}^{P}\sum\limits_{k = 1}^K{({{a}_{kp}} - {{\hat{a}}_{kp}} )}^2},
\end{equation}
where $\hat{a}_{kp}$ is an estimated abundances of the $k$th class at the $p$th pixel. RMSE between true and estimated shadow compensated reflectance has been defined as
\begin{equation}
\text{RMSE}_r =\sqrt{\frac{1}{P B}\sum\limits_{p = 1}^{P}\sum\limits_{b = 1}^B{({{r}_{bp}} - {{\hat{r}}_{bp}} )}^2},
\end{equation}
where $\hat{r}_{bp}$ is an estimated shadow compensated reflectance of the $b$th band at the $p$th pixel. The normalized reconstruction error (NRE) has been used to validate the performance of fitting. NRE has been defined as
\begin{equation}
\text{NRE} =\frac{1}{x_{max}-x_{min}}\sqrt{\frac{1}{P B}\sum\limits_{p = 1}^{P}\sum\limits_{b = 1}^B{({{x}_{bp}} - {{\hat{x}}_{bp}} )}^2},
\end{equation}
where ${x}_{bp}$ is an observed radiance or reflectance spectrum of the $b$th band at the $p$th pixel, $\hat{x}_{bp}$ is a reconstructed reflectance or radiance spectrum, $x_{max}$ is a maximum value while $x_{min}$ is a minimum value.
\begin{figure}[h]
                \centering
                \includegraphics[width=\linewidth]{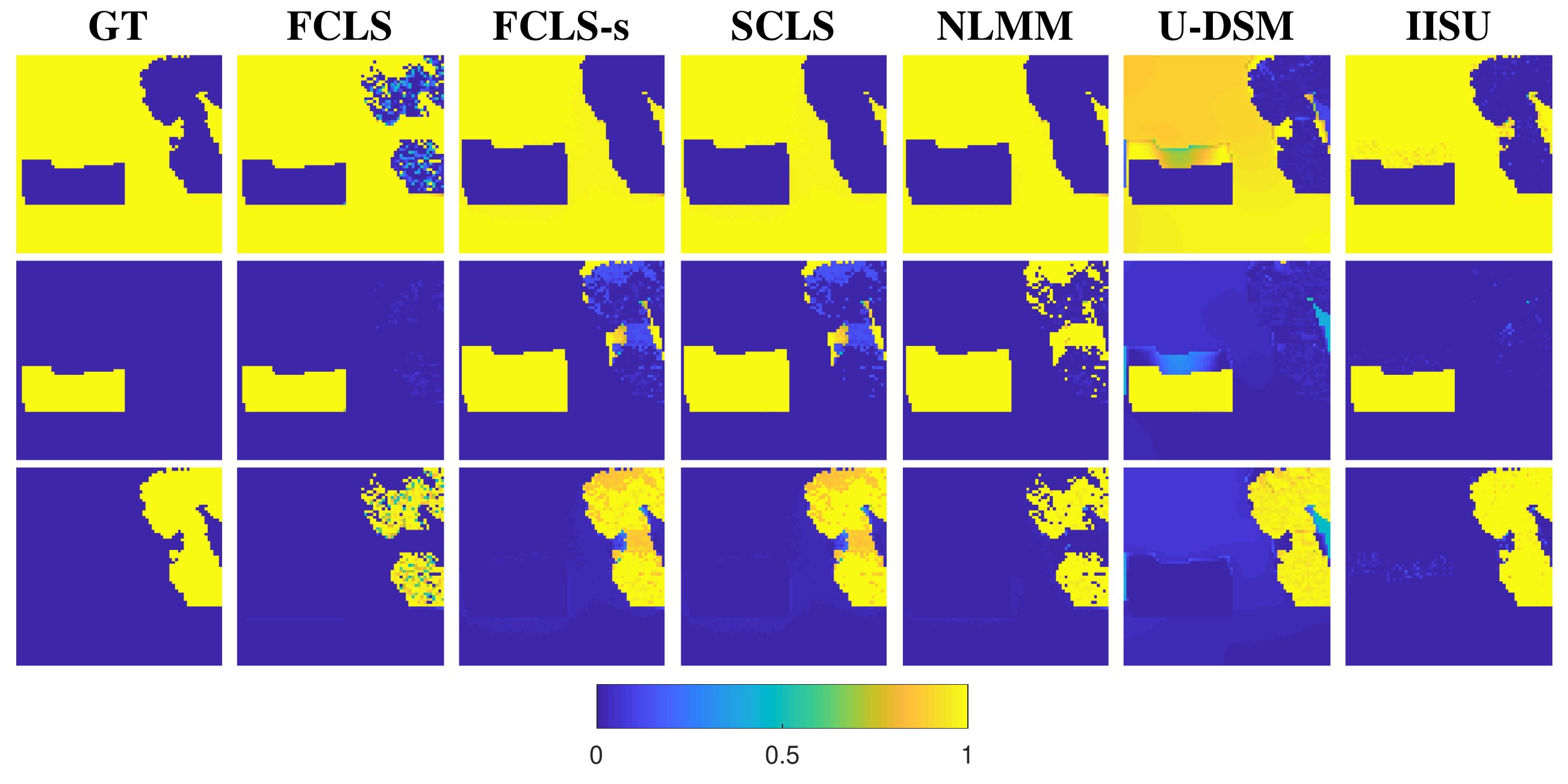}
        \caption{Ground truth and abundance maps estimated from SIM1. From top to down, the abundance maps represent a soil, a building and a tree.}\label{fig:esti_abun_sim1}
\end{figure}
\begin{figure}[h]
                \centering
                \includegraphics[width=\linewidth]{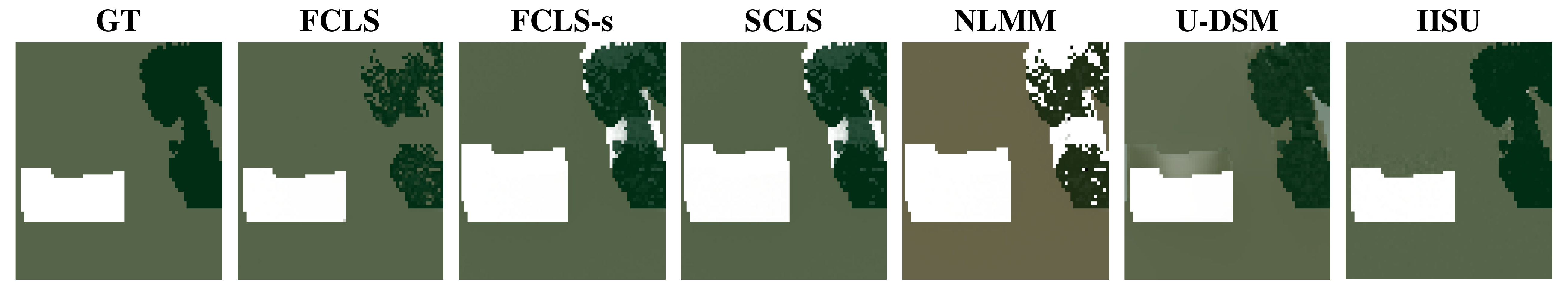}
        \caption{Ground truth and RGB imagery of shadow compensated reflectance estimated from SIM1.}\label{fig:esti_rgb_sim1}
\end{figure}
\begin{figure}[h]
                \centering
                \includegraphics[width=\linewidth]{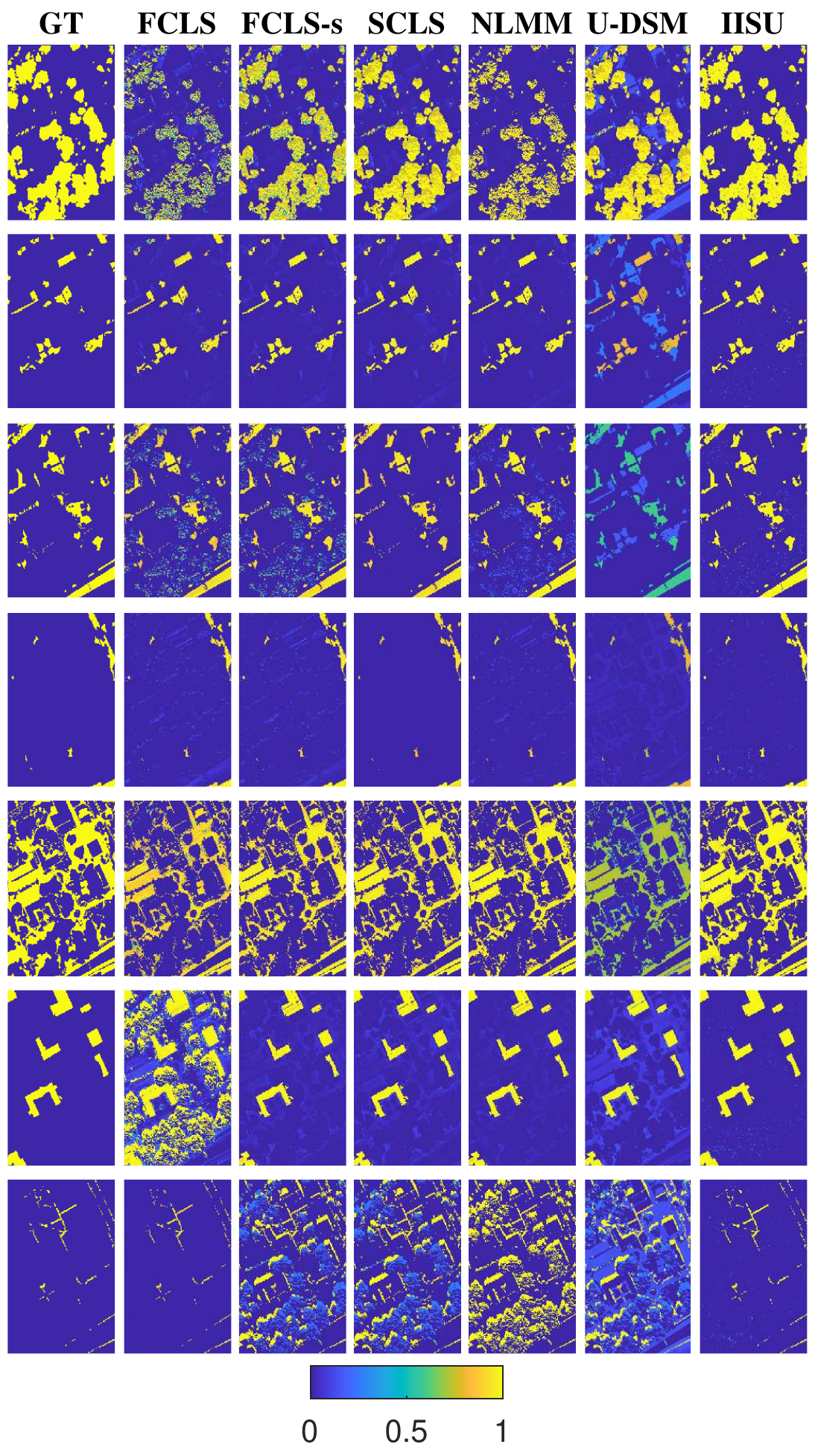}
        \caption{Ground truth and abundance maps estimated from SIM2. From top to down, the abundance maps represent trees, grass, ground surfaces, sands, roads, buildings and sidewalk.}\label{fig:esti_abun_sim2}
\end{figure}
\begin{figure}[h]
                \centering
                \includegraphics[width=\linewidth]{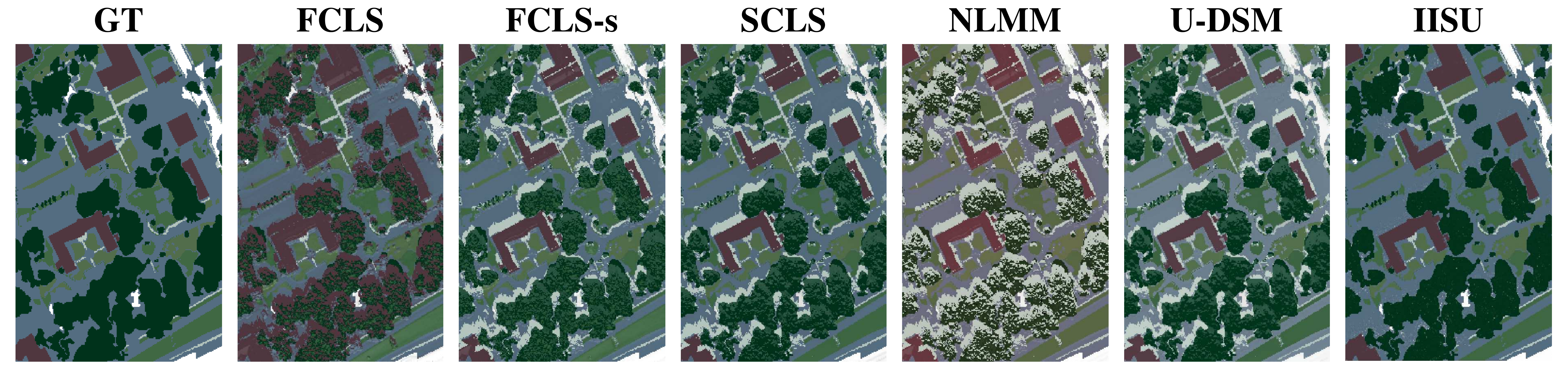}
        \caption{Ground truth and RGB imagery of shadow compensated reflectance estimated from SIM2.}\label{fig:esti_rgb_sim2}
\end{figure}
\begin{table}[h!]
\caption{RMSE$_a$ ($\times 10^{-4}$) of abundances, RMSE$_r$ ($\times 10^{-4}$) of shadow compensated reflectance and normalized RE ($\times 10^{-4}$) estimated from SIM1. The best values are shown in bold.}\label{tab:table1}
\resizebox{\columnwidth}{!}{
\begin{tabular}{ccccccccc}
\toprule							
	&FCLS	&FCLS-s	&SCLS	&NLMM	&U-DSM	&IISU	\\
\midrule							
RMSE$_a$	&16.9028	&18.7757	&18.7772	&24.8351	&7.8584	&\textbf{1.2175}	\\
RMSE$_r$	&0.7546	&1.5854	&1.586	&2.3419	&0.44	&\textbf{0.0715}	\\
NRE	&3.1888	&0.318	&0.0496	&0.3009	&0.3239	&\textbf{0.0218}	\\
\bottomrule
\end{tabular}				
}
 \end{table}
\begin{table}[h!]
\caption{RMSE$_a$ ($\times 10^{-4}$) of abundances, RMSE$_r$ ($\times 10^{-4}$) of shadow compensated reflectance and normalized RE ($\times 10^{-4}$) estimated from SIM2.}\label{tab:table2}
\resizebox{\columnwidth}{!}{
\begin{tabular}{ccccccccc}
\toprule							
	&FCLS	&FCLS-s	&SCLS	&NLMM	&U-DSM	&IISU	\\
\midrule							
RMSE$_a$	&3.8895	&2.6639	&2.4819	&3.5723	&2.7991	&\textbf{0.4236}	\\
RMSE$_r$	&0.2818	&0.1332	&0.1295	&0.2558	&0.1354	&\textbf{0.0292}	\\
NRE	&0.5043	&0.0566	&0.013	&0.0235	&0.022	&\textbf{0.0057}	\\
\bottomrule							
\end{tabular}							
}
 \end{table}
\subsection{Results}
Estimated abundances and shadow compensated reflectance derived from SIM1 showed why the existing methods failed to incorporate spectral variability caused by the shadows (Figs. \ref{fig:esti_abun_sim1} and \ref{fig:esti_rgb_sim1}). FCLS recognized all of the shadow pixels as the pixels of the soil. This showed that the projection of pixel spectra onto the space that satisfies ASC and ANC failed. The magnitude of the shadow spectra was small and the shadow spectra were outside the space that satisfies ASC and ANC. FCLS failed to correctly project the shadow spectra onto the space because FCLS did not consider the spectral variability. FCLS-s and SCLS showed qualitatively more accurate results than FCLS in the shadow pixels where the shadows were caused by the trees. However, FCLS-s and SCLS did not work well in the pixels shaded by the building. This showed that the spectral variability caused by the shadows changed not only the magnitude but also the shape of the spectra. FCLS-s and SCLS could not capture the changes in the shape of the spectra. NLMM showed the poor results. Although the bilinear endmembers can produce more flexibility, the inclusion without incorporating spectral variability may cause worse results than simple methods such as FCLS or SCLS. U-DSM produced more accurate results than FCLS, SCLS and NLMM (Table \ref{tab:table1}). This was because each spatially homogeneous regions show a different height and U-DSM could use the height information to produce accurate abundances. However, U-DSM still produced errors in the shadow pixels because it failed to incorporate the spectral variability. Unlike other methods, IISU first adjusted the endmembers according to the illumination variations of each pixel using the parameters derived from DSM. The adjusted pixel-wise endmembers led to the most accurate abundances among all methods (Table \ref{tab:table1}). IISU successfully recovered the shadow compensated reflectance in the shadow pixels (Fig. \ref{fig:esti_rgb_sim1}).

The problems of the existing methods were also observed in SIM2 (Figs. \ref{fig:esti_abun_sim2} and \ref{fig:esti_rgb_sim2}). SIM2 has more complex spatial distributions of the 7 endmembers than SIM1. FCLS performed the poorest among all methods as expected (Table \ref{tab:table2}). This was because most shadow pixels were mistakenly recognized as the spectra of the buildings. In SIM2, SCLS performed slightly better than FCLS-s. In SCLS, endmembers can be scaled to any magnitude while in FCLS-s, the scaling of an endmember is only limited by the line between the endmember and the shade endmember. This led to the superior performance of SCLS. NLMM also failed to work well in SIM2. This showed that the simple inclusion of bilinear endmembers does not necessarily improve the performance of unmixing. U-DSM performed poorly in SIM2 because there are large variations in the heights of each endmember class in the real DSM. IISU was different from U-DSM in that it first derived the physical information (visibility of the sun, sky factors or incident angles) from DSM and use the information to incorporate the spectral variability in the unmixing model. IISU produced the most accurate abundances that led to the accurate shadow compensated reflectance.
\begin{figure}[h!]
        \begin{subfigure}[b]{\columnwidth}
                \centering
                \includegraphics[width=.9\linewidth]{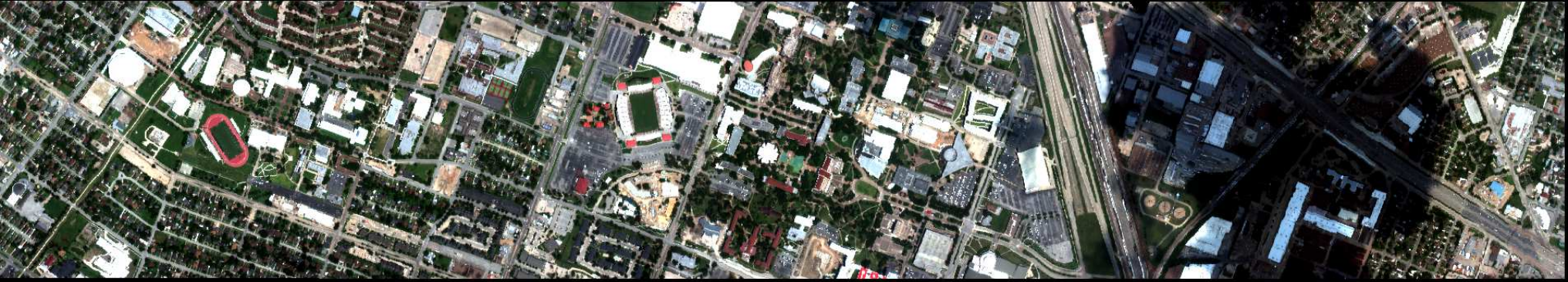}
                \caption{}
                \label{fig:rim_rgb}
        \end{subfigure}\\
        \begin{subfigure}[b]{\columnwidth}
                \centering
                \includegraphics[width=.9\linewidth]{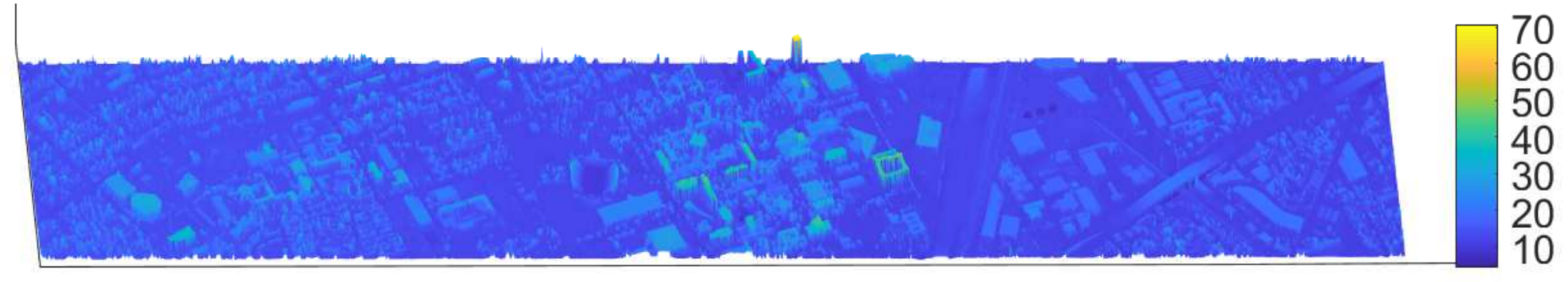}
                \caption{}
                \label{fig:rim_dsm}
        \end{subfigure}
        \caption{RIM1: (a) RGB imagery of the real HSI. (b) DSM.}\label{fig:real}
\end{figure}
\begin{figure}[h!]
        \begin{subfigure}[b]{.5\columnwidth}
                \centering
                \includegraphics[width=.9\linewidth]{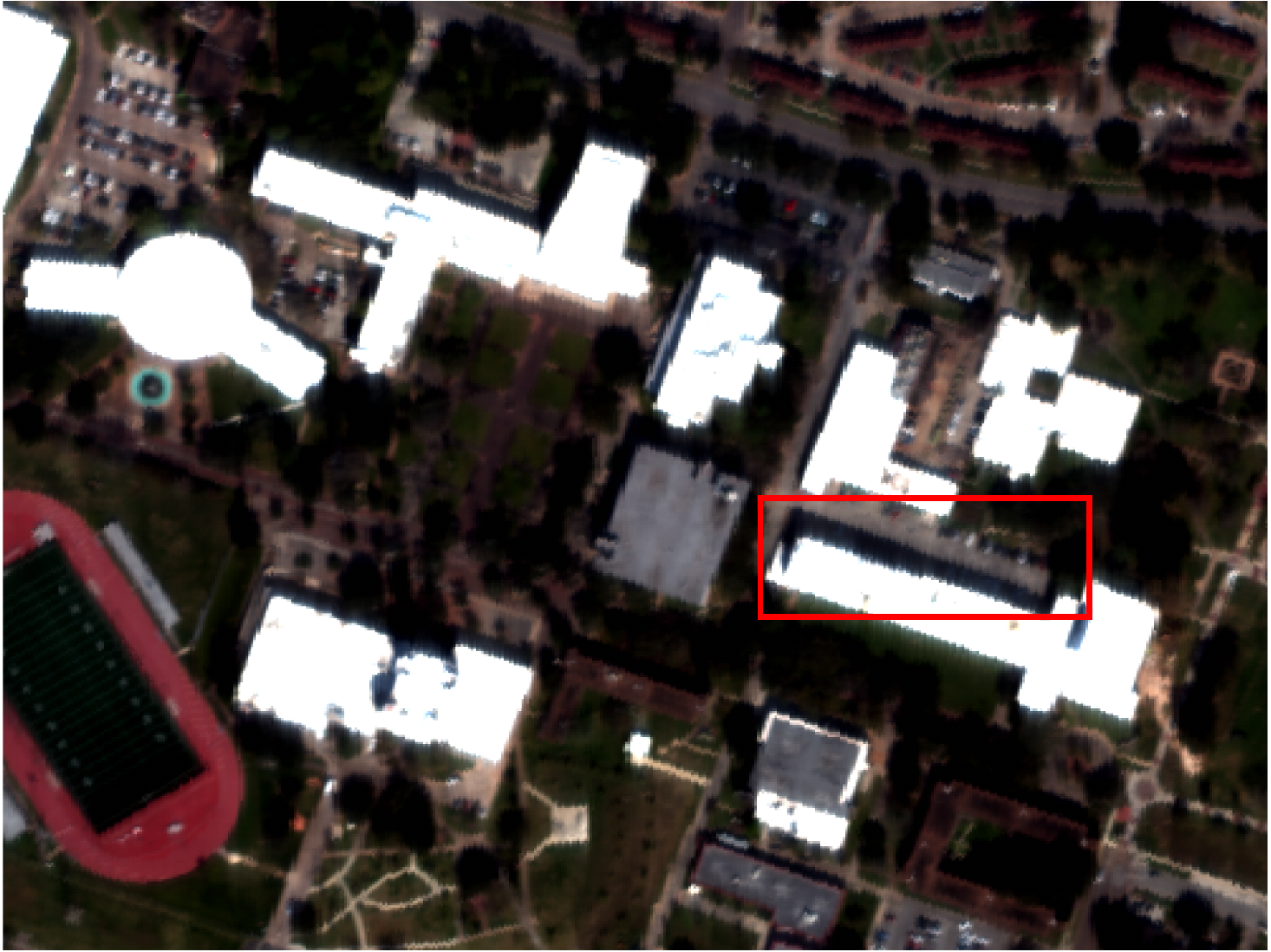}
                \caption{}
        \end{subfigure}%
        \begin{subfigure}[b]{.5\columnwidth}
                \centering
                \includegraphics[width=.9\linewidth]{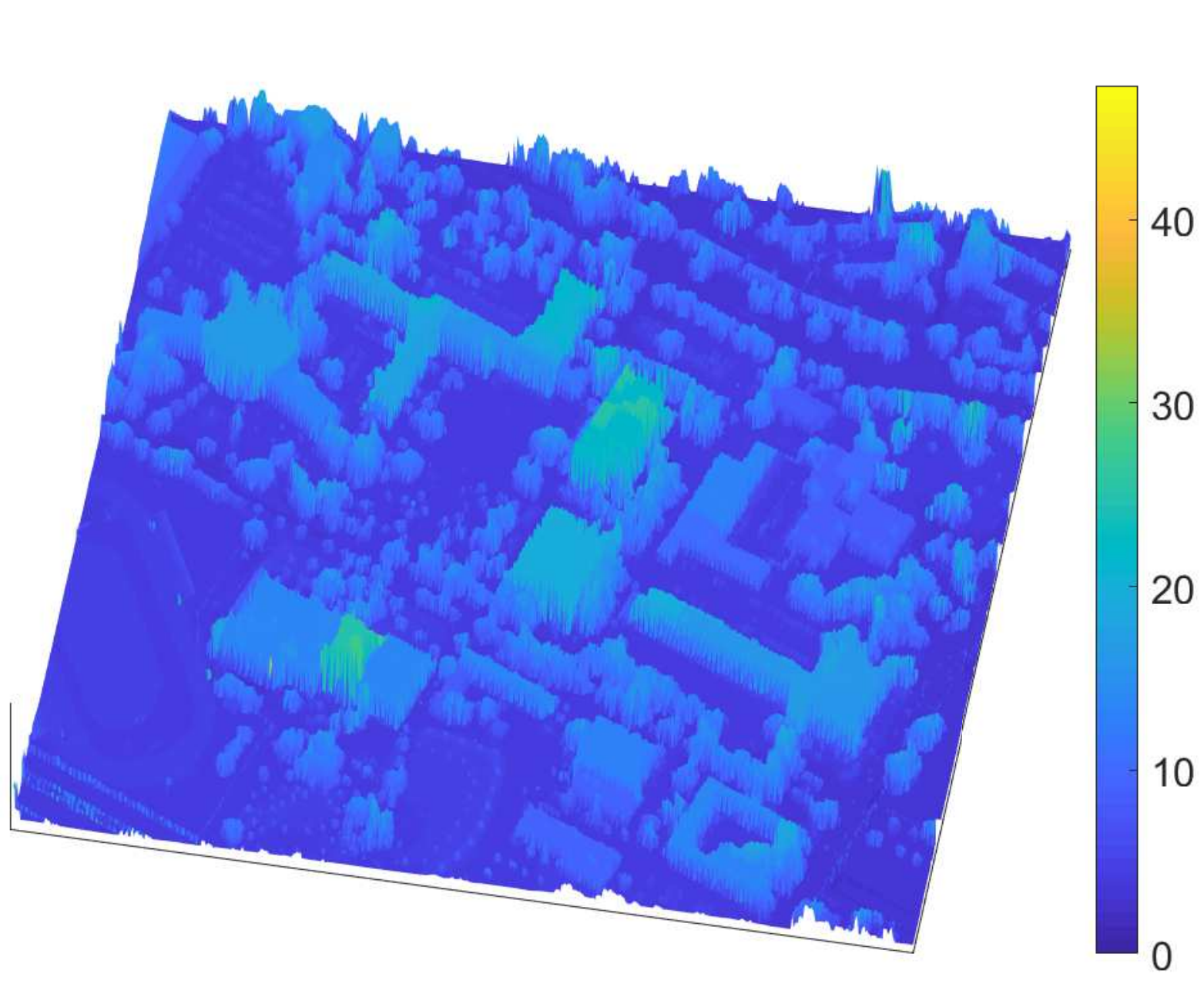}
                \caption{}
        \end{subfigure}
        \caption{\rev{RIM2: (a) RGB imagery of the real HSI. The red square represents an area affected by the large amount of the shadow. (b) DSM.}}\label{fig:real2}
\end{figure}
\begin{figure}[h]
                \centering
                \includegraphics[width=\linewidth]{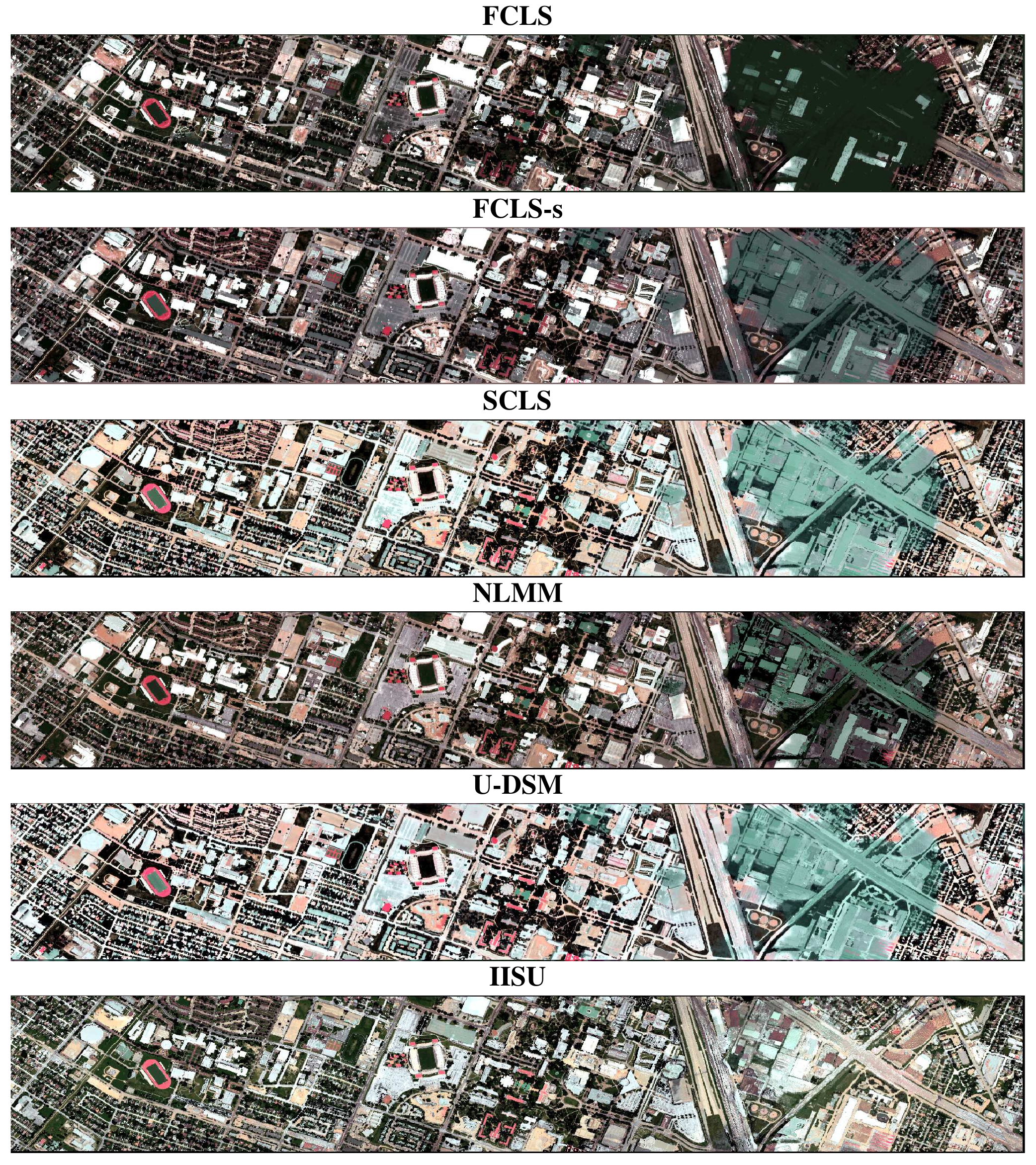}
        \caption{RIM1: RGB imagery of shadow compensated reflectance estimated by the 6 methods. All images are displayed using the $2\%$ linear stretching for fair comparison.}\label{fig:esti_rgb_rim}
\end{figure}
\begin{figure}[h]
                \centering
                \includegraphics[width=\linewidth]{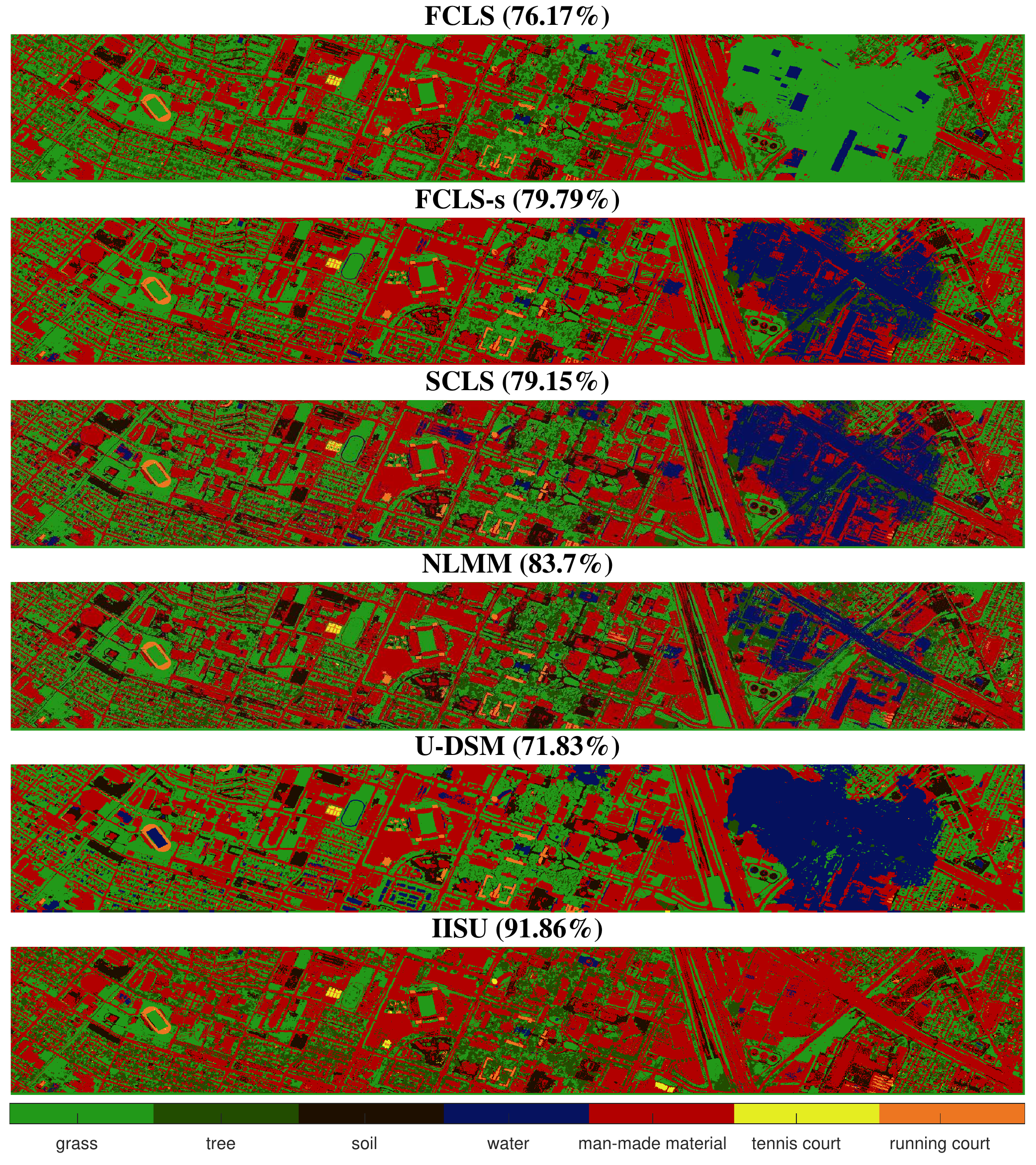}
        \caption{RIM1: classification results derived from the abundance maps of the 6 methods.}\label{fig:esti_class_rim}
\end{figure}
\begin{figure*}[h]
        \centering
        \includegraphics[width=\linewidth]{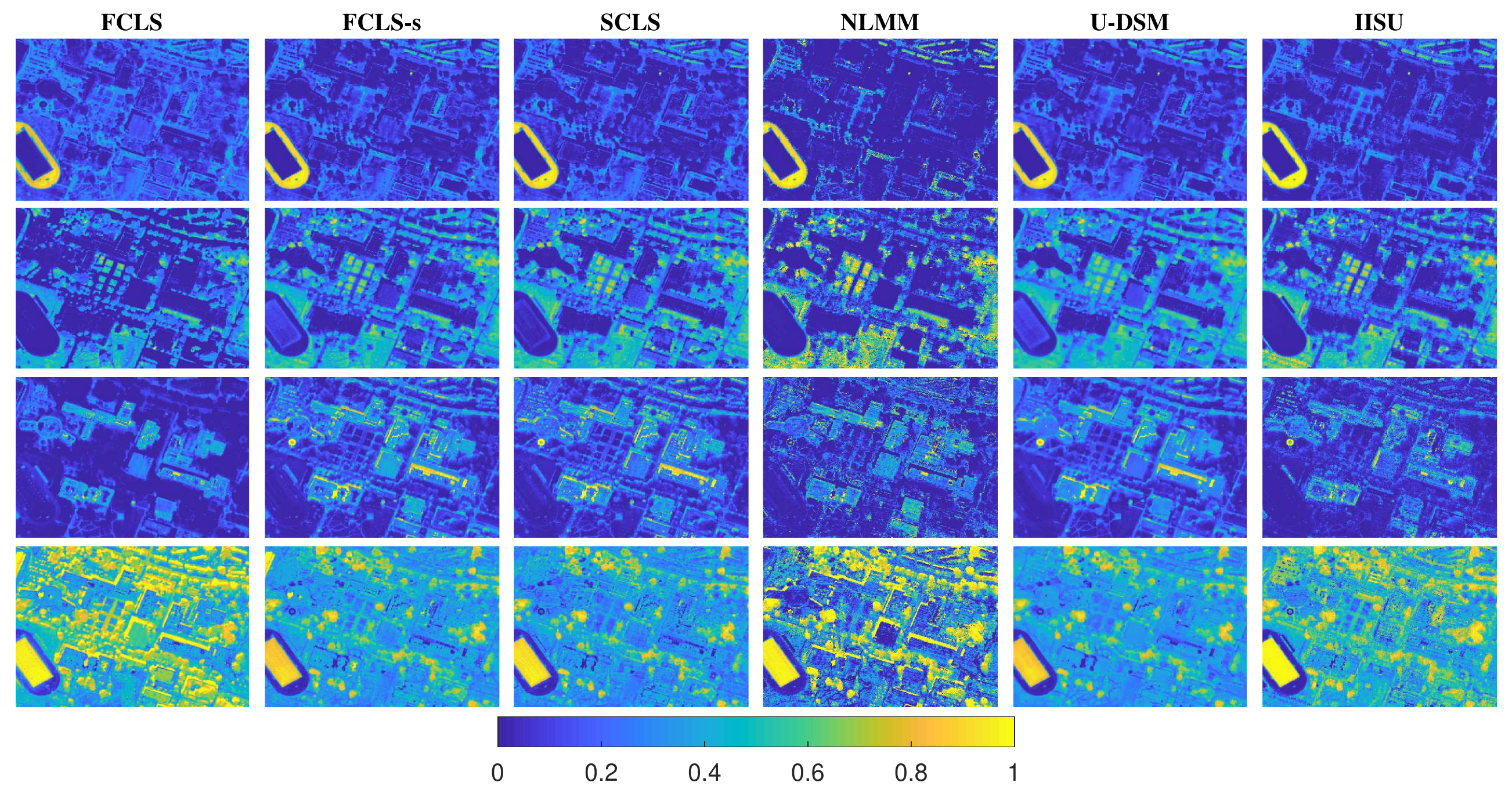}
        \caption{\rev{RIM2: abundance maps estimated by the 6 methods. From top to down, the abundance maps represent man-made material 1, vegetation 1, man-made material 2 and vegetation 2.}}\label{fig:abun_result_rim2}
\end{figure*}
\section{Experiments using real datasets}\label{section5}
\subsection{Description of real datasets}
\subsubsection{Real dataset 1 (RIM1)}
HSI and its corresponding DSM were acquired over the University of Houston campus on June 22 and June 23, 2012, respectively (Figs. \ref{fig:rim_rgb} and \ref{fig:rim_dsm})~\cite{Debes2014}. The sizes of both images are 349 x 1905 pixels. A part of HSI is affected by the cloud shadow and can be used to validate the performance of the unmixing models under the conditions where the shadow is present. Ground truth of 15 classes was available for 1200 pixels~\cite{Debes2014}. The ground sampling distance of both images is $2.5$~m. The spectral region of HSI is between 380-1050~nm and there are 144 bands. The spectral radiance was recorded in HSI.
\subsubsection{Real dataset 2 (RIM2)}
\rev{The second real dataset is available from Data Fusion Contest 2018~\cite{Xu2019}. The dataset was also acquired over the university of Houston campus on Feb 16, 2017 (Fig.~\ref{fig:real2}). However, the spatial resolution (1$m$ for HSI and 0.5$m$ for its correponding DSM) and the number of spectral bands (48 bands) are different to RIM1. RIM2 captures a higher resolution images than RIM1. In order to validate the subtle differences of abundance maps estimated by the spectral unmixing methods at a smaller region, the subset of the images was used in this study.}
\subsection{Experimental protocol}
\subsubsection{Initialization}
The radiance data of HSI in RIM1 and RIM2 were converted to the reflectance data for the methods except IISU. We followed the calibration approach proposed by~\cite{Yamaz2014}. \rev{The approach assumes that the brightest pixel in the white roof can be used as a calibration panel. The reflectance of the white roof is assumed to be $c$ in~(\ref{eq:radiance_cpanel}). The calibration approach divides each pixel radiance of HSI by that of the white roof to estimate the apparent reflectance. This approach was used in this study because the estimated reflectance spectra were confirmed to be similar to the field spectra collected by the field investigation except a shorter wavelength region~\cite{Yamaz2014}.} The short wavelength region was strongly affected by the skylight, which was inappropriate to be used for analysis. The wavelength region between 470~nm and 1050~nm was used for this study. The unmixing models in the section \ref{compared_method} were also evaluated in the datasets in the experiments.
\subsubsection{RIM1}
Fifteen endmember spectra were generated as the averages of reflectance spectra belonging to each class available from the Data Fusion Contest 2013. Although 15 endmember classes were available, spectral variability of some classes overlapped one another. The number of endmember classes present in HSI was set as 8 by merging the overlapping endmember classes. In this study, the 15 endmember spectra were firstly used for unmixing to generate 15 abundance maps. Some abundance maps belonging to a same class were merged by the summation. shadow compensated reflectance was recovered by combining the endmember spectra and abundances. Finally, the classification map was generated by selecting a class whose abundance is largest within each pixel. Because the classes of some pixels were available as ground truth, the classification maps estimated by all models were validated using the ground truth. REs of all methods were also compared. The parameter of U-DSM was empirically determined so that it can achieve the best classification result. In this HSI, the cloud shadow cannot be captured by DSM. The parameter ($v$) representing the visibility from the sun was generated as follows: First, RE between observed spectra and spectra reconstructed by FCLS was computed. Second, the cloud shadow areas were extracted and set as 1 in $v$ by applying a threshold and a morphological operator. This approach is based on the assumption that RE derived from FCLS is larger in the shadow pixels. Other physical parameters ($f$ and $\theta$) required for IISU were estimated from DSM.
\subsubsection{RIM2}
\rev{The experiment of RIM2 was designed to compare the spatial details of estimated abundance maps in an urban area, which was not considered in RIM1. Nine endmember spectra were extracted by applying VCA~\cite{Nasci2005a} to data whose shadow pixels were excluded. The shadow pixels were identified by using the visibility parameter estimated by DSM. One manually selected endmember was also added to the endmember set. The 10 endmember spectra were firstly used for unmixing to generate 10 abundance maps. Some abundances belonging to a same class were grouped by the summation. The 4 endmember classes were shown in this study. In the dataset, the ground truth of abundance estimates were not available. Thus, estimated abundance maps were qualitatively validated. REs of all methods were also computed and compared in RIM2. Finally, the parameters required by all methods have been optimized so that the parameters could achieve the best qualitative results.}
\begin{table}[ht]
\caption{RIM1: normalized RE ($\times 10^{-3}$) estimated from the real image.}\label{tab:table_rim}
\resizebox{\columnwidth}{!}{
\begin{tabular}{ccccccccc}						
\toprule							
	&FCLS	&FCLS-s	&SCLS	&NLMM	&U-DSM	&IISU	\\
\midrule
NRE	&0.2862	&0.2321	&0.0207	&0.0317	&0.0228	&\textbf{0.0148}	\\
\bottomrule							
\end{tabular}							
}
\end{table}
\begin{table}[ht]
\caption{RIM2: normalized RE ($\times 10^{-5}$) estimated from the real image.}\label{tab:table_rim2}
\resizebox{\columnwidth}{!}{
\begin{tabular}{ccccccccc}							
\toprule							
	&FCLS	&FCLS-s	&SCLS	&NLMM	&U-DSM	&IISU	\\
\midrule
NRE	&0.8901	&0.1750	&0.1600	&0.1597	&0.1783	&\textbf{0.0903}	\\
\bottomrule							
\end{tabular}
}
\end{table}
\subsection{Results}
\subsubsection{RIM1}
In order to qualitatively evaluate the shadow compensated reflectance estimated by the methods, the RGB images of the shadow compensated reflectance are shown in Fig. \ref{fig:esti_rgb_rim}. As expected, the shadow compensated reflectance recovered by FCLS showed poor results in shaded pixels. This was because FCLS simply projected the spectra of shaded pixels onto the nearest point in the space satisfying ASC and ANC without considering the spectral variability caused by the shadow. Even when a shade endmember or a scaling factor was used to suppress the spectral variability, the recovered shadow compensated reflectance was still affected by the shadow. This implies that illumination variations caused by shadows cannot be described as a simple change in the magnitude of spectra. Unlike other methods, IISU incorporates various factors from DSM and generates a flexible pixel-wise endmembers for each pixel. This led to the recovery of more realistic shadow compensated reflectance.

The classification maps were generated by selecting a class that has the largest abundance within each pixel (Fig. \ref{fig:esti_class_rim}). Note that the estimated abundance maps were shown in the supplementary document because of the limited space. FCLS recognized most of the shadow pixels as the grass because the spectra of the grass was nearest to those of the shadow pixels. SCLS, FCLS-s, NNLM and U-DSM were insensitive to the changes in the magnitude of spectra. These methods recognized that the spectral shape of the shaded pixels was similar to that of water. NNLM computed a more accurate classification result than FCLS, FCLS-s, SCLS and U-DSM. This showed that the bilinear endmembers used in NLMM could improve the accuracy of unmixing although there were still errors. U-DSM promoted the spatial homogeneity also in the shaded pixels. The height information of DSM was not effectively used when a large part of the image was obscured by the shadow. IISU performed best among all methods and also produced the lowest RE (Table \ref{tab:table_rim}).
\subsubsection{RIM2}
\rev{The abundance maps estimated by the 6 methods from RIM2 are shown in Fig.~\ref{fig:abun_result_rim2}. Each column of the figure show the 4 different  classes. The 4 endmember classes represent man-made materials~1, vegetation~1, man-made materials~2 and vegetation~2 from top to down. Unlike SIM1, SIM2 and RIM1, this dataset is not strongly affected by the cloud shadow or variable illuminations. The overall spatial patterns of the abundance maps estimated by the 6 methods are similar. This is reasonable because IISU should perform as well as other existing unmixing methods in the pixels that are not affected by the variable illuminations.  However, there were clear differences in the abundance estimates of the shaded pixels observed in Fig.~\ref{fig:real2}a. FCLS-s, SCLS and U-DSM overestimated the abundances of the man-made materials~2 in the shaded pixels. FCLS and NLMM also overestimated the abundances of vegetation~2 in the shaded pixels. IISU was less sensitive to the effects of the shadow, compared with other methods. The results show the robustness of IISU to estimate abundances in the shaded pixels as it was also demonstrated in simulated data. Also in RIM2, IISU produced the lowest RE. (Table \ref{tab:table_rim2})}
\section{Conclusion}\label{section6}
This paper proposed an unmixing model that fuses hyperspectral data and LiDAR-derived DSM. The proposed model is novel and different from existing unmixing models as follows: 1)  it generalizes and describes most of the existing unmixing models using the radiative transfer theory; 2) it naturally incorporates physical parameters from DSM in the unmixing model; 3) it opens up possibilities to directly use radiance data in the unmixing process. Experiments showed that the widely-used unmixing models does not work well when variable illuminations including shadows are present. This is because the models cannot incorporate various factors (skylight, sky factor or visibility of the sun). IISU successfully incorporated the various factors from LiDAR-derived DSM in the unmixing model and outperformed other models while using a straightforward optimization procedure. As a results, the derived shadow compensated reflectance data were less sensitive to the variable illuminations. This paper used only a simple constraint (i.e., nonnegativity) in order to show that a novel model can work well without relying on the extensive use of constraints. Future work includes the use of other constraints in order to further improve the accuracy of abundance estimates. Another direction of research is to consider spectral variability caused by other factors (e.g., physical characteristics of materials). By using the proposed model, we can explicitly discriminate the different types of spectral variability caused by variable illuminations or internal properties of materials. \rev{The proposed unmixing model can be potentially extended to incorporate other nonlinear mixing models such as intimate mixtures of minerals or tree canopies, or multilinear mixtures. The model that incorporates orthophoto-derived DSM would be of interest for future work.}




\ifCLASSOPTIONcaptionsoff
  \newpage
\fi


\bibliographystyle{IEEEtran}
\bibliography{strings_all_ref,ref_all}

\begin{thebibliography}{10}
\providecommand{\url}[1]{#1}
\csname url@samestyle\endcsname
\providecommand{\newblock}{\relax}
\providecommand{\bibinfo}[2]{#2}
\providecommand{\BIBentrySTDinterwordspacing}{\spaceskip=0pt\relax}
\providecommand{\BIBentryALTinterwordstretchfactor}{4}
\providecommand{\BIBentryALTinterwordspacing}{\spaceskip=\fontdimen2\font plus
\BIBentryALTinterwordstretchfactor\fontdimen3\font minus
  \fontdimen4\font\relax}
\providecommand{\BIBforeignlanguage}[2]{{%
\expandafter\ifx\csname l@#1\endcsname\relax
\typeout{** WARNING: IEEEtran.bst: No hyphenation pattern has been}%
\typeout{** loaded for the language `#1'. Using the pattern for}%
\typeout{** the default language instead.}%
\else
\language=\csname l@#1\endcsname
\fi
#2}}
\providecommand{\BIBdecl}{\relax}
\BIBdecl

\bibitem{Griff2003}
M.~K. Griffin and H.-h.~K. Burke, ``Compensation of {hyperspectral} {data} for
  {atmospheric} {effects},'' in \emph{Lincolm {Laboratory} {Journal}}, ser. 1,
  vol.~14, 2003, pp. 29--54.

\bibitem{Kesha2002}
N.~Keshava and J.~F. Mustard, ``Spectral unmixing,'' \emph{IEEE Signal Process.
  Mag.}, vol.~19, no.~1, pp. 44--57, 2002.

\bibitem{Biouc2012}
J.~M. Bioucas-Dias, A.~Plaza, N.~Dobigeon, M.~Parente, D.~Qian, P.~Gader, and
  J.~Chanussot, ``Hyperspectral {unmixing} {overview}: {geometrical},
  {statistical}, and {sparse} {regression}-{based} {approaches},'' \emph{IEEE
  J. Sel. Topics Appl. Earth Observations Remote Sens.}, vol.~5, no.~2, pp.
  354--379, 2012.

\bibitem{Murph2012}
R.~J. Murphy, S.~T. Monteiro, and S.~Schneider, ``Evaluating {classification}
  {techniques} for {mapping} {vertical} {geology} {using} {field}-{based}
  {hyperspectral} {sensors},'' \emph{IEEE Trans. Geosci. Remote Sens.},
  vol.~50, no.~8, pp. 3066--3080, 2012.

\bibitem{Uezat2016}
T.~Uezato, R.~J. Murphy, A.~Melkumyan, and A.~Chlingaryan, ``A {novel}
  {endmember} {bundle} {extraction} and {clustering} {approach} for {capturing}
  {spectral} {variability} within {endmember} {classes},'' \emph{IEEE Trans.
  Geosci. Remote Sens.}, vol.~54, no.~11, pp. 6712--6731, 2016.

\bibitem{Somer2011}
B.~Somers, G.~P. Asner, L.~Tits, and P.~Coppin, ``Endmember variability in
  {spectral} {mixture} {analysis}: {a} review,'' \emph{Remote Sens.
  Environment}, vol. 115, no.~7, pp. 1603--1616, 2011.

\bibitem{Zare2014a}
A.~Zare and K.~C. Ho, ``Endmember {variability} in {hyperspectral} {analysis}:
  {addressing} {spectral} {variability} {during} {spectral} {unmixing},''
  \emph{IEEE Signal Process. Mag.}, vol.~31, no.~1, pp. 95--104, 2014.

\bibitem{Rober1998}
D.~A. Roberts, M.~Gardner, R.~Church, S.~Ustin, G.~Scheer, and R.~O. Green,
  ``Mapping {chaparral} in the {santa} {monica} {mountains} {using} {multiple}
  {endmember} {spectral} {mixture} {models},'' \emph{Remote Sens. Environment},
  vol.~65, no.~3, pp. 267--279, 1998.

\bibitem{Nasci2005a}
J.~M.~P. Nascimento and J.~M. Bioucas~Dias, ``Vertex component analysis: a fast
  algorithm to unmix hyperspectral data,'' \emph{IEEE Trans. Geosci. Remote
  Sens.}, vol.~43, no.~4, pp. 898--910, 2005.

\bibitem{Winte1999}
M.~E. Winter, ``N-{FINDR}: an algorithm for fast autonomous spectral end-member
  determination in hyperspectral data,'' in \emph{Proceedings of {SPIE}}, vol.
  3753, 1999, pp. 266--275.

\bibitem{Uezat2016a}
T.~Uezato, R.~J. Murphy, A.~Melkumyan, and A.~Chlingaryan, ``A {novel}
  {spectral} {unmixing} {method} {incorporating} {spectral} {variability}
  {within} {endmember} {classes},'' \emph{IEEE Trans. Geosci. Remote Sens.},
  vol.~54, no.~5, pp. 2812--2831, 2016.

\bibitem{Fitzg2005}
G.~J. Fitzgerald, P.~J. Pinter~Jr, D.~J. Hunsaker, and T.~R. Clarke, ``Multiple
  shadow fractions in spectral mixture analysis of a cotton canopy,''
  \emph{Remote Sens. Environment}, vol.~97, no.~4, pp. 526--539, 2005.

\bibitem{Drume2016}
L.~Drumetz, M.~A. Veganzones, S.~Henrot, R.~Phlypo, J.~Chanussot, and
  C.~Jutten, ``Blind {hyperspectral} {unmixing} {using} an {extended} {linear}
  {mixing} {model} to {address} {spectral} {variability},'' \emph{IEEE Trans.
  Image Process.}, vol.~25, no.~8, pp. 3890--3905, 2016.

\bibitem{Thouv2016}
P.~A. Thouvenin, N.~Dobigeon, and J.~Y. Tourneret, ``Hyperspectral {unmixing}
  {with} {spectral} {variability} {using} a {perturbed} {linear} {mixing}
  {model},'' \emph{IEEE Trans. Signal Process.}, vol.~64, no.~2, pp. 525--538,
  2016.

\bibitem{Uezat2016b}
T.~Uezato, R.~J. Murphy, A.~Melkumyan, and A.~Chlingaryan, ``Incorporating
  {spatial} {information} and {endmember} {variability} {into} {unmixing}
  {analyses} to {improve} {abundance} {estimates},'' \emph{IEEE Trans. Image
  Process.}, vol.~25, no.~12, pp. 5563--5575, 2016.

\bibitem{Ramak2015}
R.~Ramakrishnan, J.~Nieto, and S.~Scheding, ``Shadow compensation for outdoor
  perception,'' in \emph{{IEEE} {International} {Conference} on {Robotics} and
  {Automation} ({ICRA})}, 2015, pp. 4835--4842.

\bibitem{Frima2011}
O.~Friman, G.~Tolt, and J.~Ahlberg, ``Illumination and shadow compensation of
  hyperspectral images using a digital surface model and non-linear least
  squares estimation,'' in \emph{Image and {Signal} {Processing} for {Remote}
  {Sensing} {XVII}}, vol. 8180.\hskip 1em plus 0.5em minus 0.4em\relax
  International Society for Optics and Photonics, 2011, p. 81800Q.

\bibitem{Dobig2014}
N.~Dobigeon, J.~Y. Tourneret, C.~Richard, J.~C.~M. Bermudez, S.~McLaughlin, and
  A.~O. Hero, ``Nonlinear {unmixing} of {hyperspectral} {images}: {models} and
  {algorithms},'' \emph{IEEE Signal Process. Mag.}, vol.~31, no.~1, pp. 82--94,
  2014.

\bibitem{Heyle2014}
R.~Heylen, M.~Parente, and P.~Gader, ``A {review} of {nonlinear}
  {hyperspectral} {unmixing} {methods},'' \emph{IEEE J. Sel. Topics Appl. Earth
  Observations Remote Sens.}, vol.~7, no.~6, pp. 1844--1868, 2014.

\bibitem{Adeli2013}
K.~Adeline, M.~Chen, X.~Briottet, S.~Pang, and N.~Paparoditis, ``Shadow
  detection in very high spatial resolution aerial images: A comparative
  study,'' \emph{ISPRS Journal of Photogrammetry and Remote Sensing}, vol.~80,
  pp. 21--38, 2013.

\bibitem{Megan2014a}
I.~Meganem, P.~Deliot, X.~Briottet, Y.~Deville, and S.~Hosseini,
  ``Linear-{quadratic} {mixing} {model} for {reflectances} in {urban}
  {environments},'' \emph{IEEE Trans. Geosci. Remote Sens.}, vol.~52, no.~1,
  pp. 544--558, 2014.

\bibitem{Brell2017}
M.~Brell, K.~Segl, L.~Guanter, and B.~Bookhagen, ``Hyperspectral and lidar
  intensity data fusion: {A} framework for the rigorous correction of
  illumination, anisotropic effects, and cross calibration,'' \emph{IEEE Trans.
  Geosci. Remote Sens.}, vol.~55, no.~5, pp. 2799--2810, 2017.

\bibitem{Ni2014}
L.~Ni, L.~Gao, S.~Li, J.~Li, and B.~Zhang, ``Edge-constrained {Markov} random
  field classification by integrating hyperspectral image with {LiDAR} data
  over urban areas,'' \emph{Journal of Applied Remote Sensing}, vol.~8, no.~1,
  pp. 085\,089--085\,089, 2014.

\bibitem{Uezat2018}
T.~Uezato, M.~Fauvel, and N.~Dobigeon, ``Hyperspectral image unmixing with
  {LiDAR} data-aided spatial regularization,'' \emph{IEEE Trans. Geosci. Remote
  Sens.}, vol.~56, no.~7, pp. 4098--4108, 2018.

\bibitem{Feng2003}
J.~Feng, B.~Rivard, and A.~Sanchez, A.nchez-Azofeifa, ``The topographic
  normalization of hyperspectral data: implications for the selection of
  spectral end members and lithologic mapping,'' \emph{Remote Sens.
  Environment}, vol.~85, no.~2, pp. 221--231, 2003.

\bibitem{Cooley2002}
T.~Cooley, G.~P. Anderson, G.~W. Felde, M.~L. Hoke, A.~J. Ratkowski, J.~H.
  Chetwynd, J.~A. Gardner, S.~M. Adler-Golden, M.~W. Matthew, A.~Berk
  \emph{et~al.}, ``Flaash, a modtran4-based atmospheric correction algorithm,
  its application and validation,'' in \emph{Proc. IEEE Int. Conf. Geosci.
  Remote Sens. (IGARSS)}, vol.~3.\hskip 1em plus 0.5em minus 0.4em\relax IEEE,
  2002, pp. 1414--1418.

\bibitem{Richter2004}
R.~Richter and R.~S.~D. Center, ``Atcor: Atmospheric and topographic
  correction,'' \emph{German Aerospace Center, Mars: Oberpfaffenhofen,
  Germany}, 2004.

\bibitem{Schlaepfer2018}
D.~Schl{\"a}pfer, A.~Hueni, and R.~Richter, ``Cast shadow detection to quantify
  the aerosol optical thickness for atmospheric correction of high spatial
  resolution optical imagery,'' \emph{Remote Sensing}, vol.~10, no.~2, p. 200,
  2018.

\bibitem{Altma2012}
Y.~Altmann, A.~Halimi, N.~Dobigeon, and J.~Y. Tourneret, ``Supervised
  {nonlinear} {spectral} {unmixing} {using} a {postnonlinear} {mixing} {model}
  for {hyperspectral} {imagery},'' \emph{IEEE Trans. Image Process.}, vol.~21,
  no.~6, pp. 3017--3025, 2012.

\bibitem{Yokoy2014}
N.~Yokoya, J.~Chanussot, and A.~Iwasaki, ``Nonlinear {unmixing} of
  {hyperspectral} {data} {using} {semi}-{nonnegative} {matrix}
  {factorization},'' \emph{IEEE Trans. Geosci. Remote Sens.}, vol.~52, no.~2,
  pp. 1430--1437, 2014.

\bibitem{Megan2014}
I.~Meganem, Y.~Deville, S.~Hosseini, D.~P, {x00E}, {liot}, and X.~Briottet,
  ``Linear-{quadratic} {blind} {source} {separation} {using} {NMF} to {unmix}
  {urban} {hyperspectral} {images},'' \emph{IEEE Trans. Signal Process.},
  vol.~62, no.~7, pp. 1822--1833, 2014.

\bibitem{qu2014}
Q.~Qu, N.~M. Nasrabadi, and T.~D. Tran, ``Abundance {estimation} for {bilinear}
  {mixture} {models} via {joint} {sparse} and {low}-{rank} {representation},''
  \emph{IEEE Trans. Geosci. Remote Sens.}, vol.~52, no.~7, pp. 4404--4423,
  2014.

\bibitem{Liu2017}
Y.~Liu, J.~Bioucas-Dias, J.~Li, and A.~Plaza, ``Hyperspectral cloud shadow
  removal based on linear unmixing,'' in \emph{{Proc. IEEE Int. Conf. Geosci.
  Remote Sens. (IGARSS)}}, July 2017, pp. 1000--1003.

\bibitem{Hong2019}
D.~Hong, N.~Yokoya, J.~Chanussot, and X.~X. Zhu, ``An {augmented} {linear}
  {mixing} {model} to {address} {spectral} {variability} for {hyperspectral}
  {unmixing},'' \emph{IEEE Trans. Image Process.}, vol.~28, no.~4, pp.
  1923--1938, April 2019.

\bibitem{Imbir2017}
T.~Imbiriba, R.~A. Borsoi, and J.~C.~M. Bermudez, ``Generalized linear mixing
  model accounting for endmember variability,'' \emph{arXiv preprint
  arXiv:1710.07723}, 2017.

\bibitem{Uezat2019}
T.~Uezato, M.~Fauvel, and N.~Dobigeon, ``Hyperspectral unmixing with spectral
  variability using adaptive bundles and double sparsity,'' \emph{IEEE Trans.
  Geosci. Remote Sens.}, vol.~57, no.~6, 2019.

\bibitem{Drume2019}
L.~Drumetz, T.~R. Meyer, J.~Chanussot, A.~L. Bertozzi, and C.~Jutten,
  ``Hyperspectral {image} {unmixing} with {endmember} {bundles} and {group}
  {sparsity} {inducing} {mixed} {norms},'' \emph{IEEE Trans. Image Process.},
  2019.

\bibitem{Zhou2018}
Y.~Zhou, A.~Rangarajan, and P.~D. Gader, ``A {Gaussian} mixture model
  representation of endmember variability in hyperspectral unmixing,''
  \emph{IEEE Trans. Image Process.}, vol.~27, no.~5, pp. 2242--2256, 2018.

\bibitem{Halim2016}
A.~Halimi, P.~Honeine, and J.~M. Bioucas-Dias, ``Hyperspectral unmixing in
  presence of endmember variability, nonlinearity, or mismodeling effects,''
  \emph{IEEE Trans. Image Process.}, vol.~25, no.~10, pp. 4565--4579, 2016.

\bibitem{Heinz2001}
D.~C. Heinz and I.~C. Chein, ``Fully constrained least squares linear spectral
  mixture analysis method for material quantification in hyperspectral
  imagery,'' \emph{IEEE Trans. Geosci. Remote Sens.}, vol.~39, no.~3, pp.
  529--545, 2001.

\bibitem{Boyd2011}
S.~Boyd, N.~Parikh, E.~Chu, B.~Peleato, and J.~Eckstein, ``Distributed
  optimization and statistical learning via the alternating direction method of
  multipliers,'' \emph{Foundations and Trends® in Machine Learning}, vol.~3,
  no.~1, pp. 1--122, 2011.

\bibitem{Sun2014}
D.~L. Sun and C.~Fevotte, ``Alternating direction method of multipliers for
  non-negative matrix factorization with the beta-divergence,'' in \emph{2014
  IEEE international conference on acoustics, speech and signal processing
  (ICASSP)}.\hskip 1em plus 0.5em minus 0.4em\relax IEEE, 2014, pp. 6201--6205.

\bibitem{Gader2013}
P.~Gader, A.~Zare, R.~Close, J.~Aitken, and G.~Tuell, ``{M}uufl gulfport
  hyperspectral and lidar airborne data set. {U}niv. {F}lorida, {G}ainesville,
  {FL},'' USA, Tech. Rep. REP-2013-570, Tech. Rep., 2013.

\bibitem{Gueym1995}
C.~Gueymard, ``Smarts2 a simple model of the atmospheric radiative transfer of
  sunshine: algorithms and performance assessment. florida solar energy center
  rep,'' FSEC-PF-270-95, Tech. Rep., 1995.

\bibitem{du2017}
X.~Du and A.~Zare, ``Technical report: Scene label ground truth map for muufl
  gulfport data set,'' \emph{University of Florida, Gainesville, FL, Tech. Rep.
  20170417}, 2017.

\bibitem{Nasci2009}
J.~M.~P. Nascimento and J.~M. Bioucas-Dias, ``Nonlinear mixture model for
  hyperspectral unmixing,'' in \emph{Proceeding, {SPIE}}, vol. 7477, 2009, pp.
  74\,770I--74\,770I--8.

\bibitem{Debes2014}
C.~Debes, A.~Merentitis, R.~Heremans, J.~Hahn, N.~Frangiadakis, T.~van
  Kasteren, W.~Liao, R.~Bellens, A.~Pi{\v{z}}urica, S.~Gautama \emph{et~al.},
  ``Hyperspectral and lidar data fusion: Outcome of the 2013 grss data fusion
  contest,'' \emph{IEEE J. Sel. Topics Appl. Earth Observations Remote Sens.},
  vol.~7, no.~6, pp. 2405--2418, 2014.

\bibitem{Xu2019}
Y.~{Xu}, B.~{Du}, L.~{Zhang}, D.~{Cerra}, M.~{Pato}, E.~{Carmona}, S.~{Prasad},
  N.~{Yokoya}, R.~{Hänsch}, and B.~{Le Saux}, ``Advanced multi-sensor optical
  remote sensing for urban land use and land cover classification: Outcome of
  the 2018 {IEEE} grss data fusion contest,'' \emph{IEEE Journal of Selected
  Topics in Applied Earth Observations and Remote Sensing}, vol.~12, no.~6, pp.
  1709--1724, June 2019.

\bibitem{Yamaz2014}
F.~Yamazaki, K.~Hara, and W.~Liu, ``Urban land-cover classification based on
  airborne hyperspectral data and field observation,'' in \emph{Image and
  Signal Processing for Remote Sensing XX}, vol. 9244.\hskip 1em plus 0.5em
  minus 0.4em\relax International Society for Optics and Photonics, 2014, p.
  92440P.

\end{thebibliography}
\end{document}